%% file: ijcai25.tex
\documentclass{article}
\usepackage{ijcai25}

\usepackage{times}
\usepackage{soul}
\usepackage{url}
\usepackage[hidelinks]{hyperref}
\usepackage[utf8]{inputenc}
\usepackage[small]{caption}
\usepackage{graphicx}
\usepackage{amsmath}
\usepackage{amsthm}
\usepackage{booktabs}
\usepackage{algorithm}
\usepackage{algorithmic}
\usepackage[switch]{lineno}
\usepackage[most]{tcolorbox}

\usepackage{kotex}
\usepackage{amsmath}
\usepackage{amssymb}
\usepackage{rotating}
\usepackage{multibib}
\usepackage{multirow, multicol, xcolor}
\usepackage{mathtools}

\usepackage{hyperref}
\usepackage{url}
\usepackage{array,multirow,graphicx}
\usepackage{subcaption}

\usepackage{kotex}
\usepackage{pifont}

\usepackage{lscape}

\definecolor{mygrey}{rgb}{0.9,0.9,0.9}
\title{TiVaT: A Transformer with a Single Unified Mechanism for Capturing Asynchronous Dependencies in Multivariate Time Series Forecasting}
\author{
Junwoo Ha$^1$
\and
Hyukjae Kwon$^2$\and
Sungsoo Kim$^2$\and
Kisu Lee$^2$\and
Seungjae Park$^1$\And
Ha Young Kim$^{2,}$\thanks{Corresponding author}\\
\affiliations
$^1$Department of AI, Yonsei University, Seoul, South Korea\\
$^2$Graduate School of Information, Yonsei University, Seoul, South Korea\\
\emails
\{gkwnsdn0402,kwonhj1015,kss8421,kisu0928,seungjae.park,hayoung.kim\}@yonsei.ac.kr
}

\begin{document}

\maketitle
\begin{abstract}
Multivariate time series (MTS) forecasting is vital across various domains but remains challenging due to the need to simultaneously model temporal and inter-variate dependencies.
Existing channel-dependent models, where Transformer-based models dominate, process these dependencies separately, limiting their capacity to capture complex interactions such as lead-lag dynamics. 
To address this issue, we propose TiVaT (Time-variate Transformer), a novel architecture incorporating a single unified module, a Joint-Axis (JA) attention module, that concurrently processes temporal and variate modeling. 
The JA attention module dynamically selects relevant features to particularly capture asynchronous interactions.
In addition, we introduce distance-aware time-variate sampling in the JA attention, a novel mechanism that extracts significant patterns through a learned 2D embedding space while reducing noise. 
Extensive experiments demonstrate TiVaT’s overall performance across diverse datasets, particularly excelling in scenarios with intricate asynchronous dependencies.
\end{abstract}
% clean ver
% \begin{abstract}
% Multivariate time series (MTS) forecasting is vital across various domains but remains challenging due to the need to simultaneously model temporal and inter-variate dependencies.
% Existing channel-dependent models, where Transformer-based models dominate, process these dependencies separately, limiting their capacity to capture complex interactions such as lead-lag dynamics. 
% To address this issue, we propose TiVaT (Time-variate Transformer), a novel architecture incorporating a single unified module, a Joint-Axis (JA) attention module, that concurrently processes temporal and variate modeling. 
% The JA attention module dynamically selects relevant features to particularly capture asynchronous interactions.
% In addition, we introduce distance-aware time-variate sampling in the JA attention, a novel mechanism that extracts significant patterns through a learned 2D embedding space while reducing noise. 
% Extensive experiments demonstrate TiVaT’s overall performance across diverse datasets, particularly excelling in scenarios with intricate asynchronous dependencies.
% \end{abstract}

\input{1_Introduction}

\input{2_Related_Works}

\input{3_Methodology}

\input{4_Experiments}

\input{5_Conclusion}

\section*{Ethical Statement}

There are no ethical issues.

% \section*{Acknowledgments}

% This work was conducted without any specific funding or external support. The authors have no acknowledgments to declare.

\bibliographystyle{named}
\bibliography{ijcai25}

\newpage
\input{6_Appendix}

\end{document}

%% file: 1_Introduction.tex
\begin{figure*}[t]
\begin{subfigure}{0.49\linewidth}
    \centering
    \includegraphics[width=\textwidth]{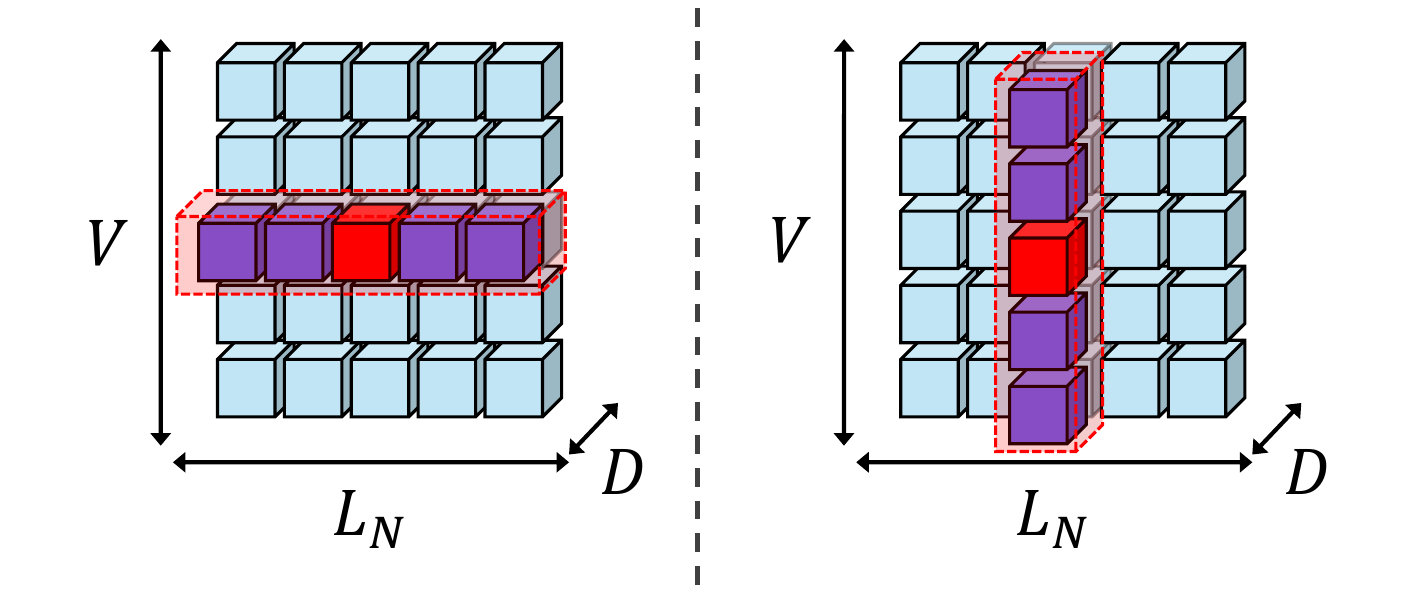}
    \caption{Conventional CD Attention}
    \label{fig1b}
\end{subfigure}
\hfill
\begin{subfigure}{0.25\linewidth}
    \centering
    \includegraphics[width=\textwidth]{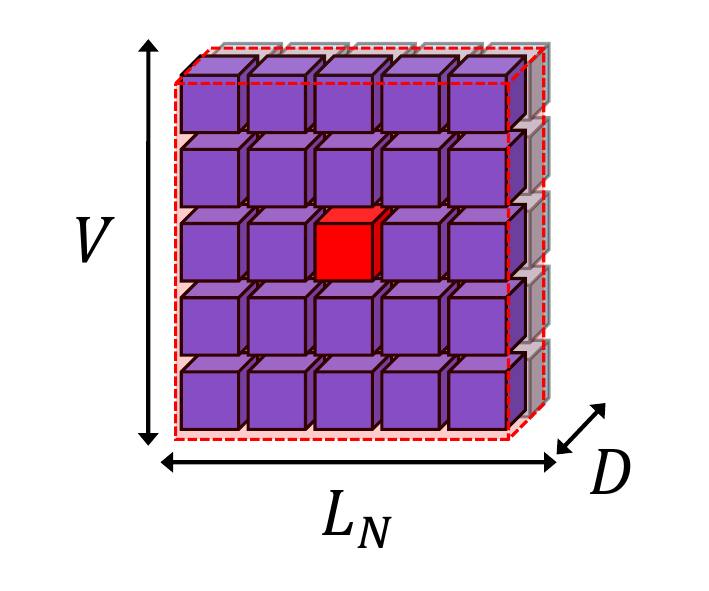}
    \caption{Full Attention}
    \label{fig1a}
\end{subfigure}
\hfill
\begin{subfigure}{0.25\linewidth}
    \centering
    \includegraphics[width=\textwidth]{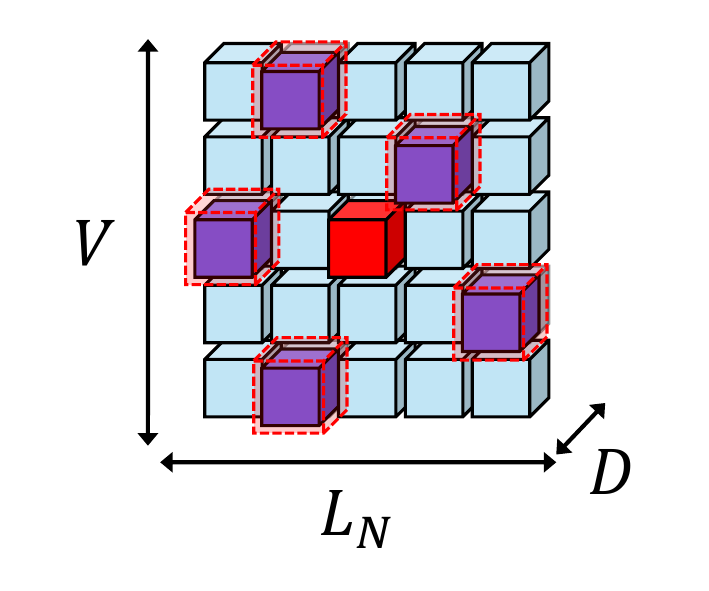}
    \caption{Joint-Axis Attention (Ours)}
    \label{fig1c}
\end{subfigure}

\hfill

\caption{Comparison of Transformer-based CD models' attention mechanisms, \(L_N\) represents the patched time axis, \(V\) denotes the variate axis, and \(D\) indicates the dimensional space.
The red box represents the feature serving as the query in attention, while the purple box represents the features serving as key-value pairs.
}
\end{figure*}

\section{Introduction}
Multivariate time series (MTS) forecasting plays a pivotal role in real-world applications such as finance (e.g., stock price prediction)~\cite{lu2024trnn}, weather modeling~\cite{angryk2020multivariate,nguyen2023climax}, traffic management~\cite{yin2016forecasting,jin2023trafformer}, and energy demand prediction~\cite{yuan2023attention}.
While early deep learning architectures like multilayer perceptrons (MLPs)~\cite{oreshkin2019n,zeng2023transformers,challu2023nhits,li2023mts}, recurrent neural networks (RNNs)~\cite{salinas2020deepar,lai2018modeling,qin2017dual}, convolutional neural networks (CNNs)~\cite{luo2024moderntcn,wu2023timesnet,wang2023micn}, and Transformers~\cite{zhou2021informer,wu2021autoformer,zhou2022fedformer,liu2022non} have made remarkable advancements, effectively capturing the intricate temporal patterns and inter-variate relationships in MTS data.

MTS forecasting models can be broadly categorized into Channel-Independent (CI) models, which treat variates independently, and Channel-Dependent (CD) models, which capture relationships between variates.
CI models~\cite{zeng2023transformers,nie2023a,wang2024timemixer} facilitate the mitigation of overfitting and noise but fail to consider inter-variate dependencies, limiting prediction accuracy~\cite{han2024capacity}.
In contrast, CD models~\cite{wang2024timexer,liu2024itransformer,yu2023dsformer,zhang2023crossformer} are designed to capture complex inter-variate interactions and long-range dependencies, and they are primarily implemented using Transformer-based architectures~\cite{wang2024timexer,liu2024itransformer,yu2023dsformer,zhang2023crossformer} that leverage the self-attention mechanism.

As illustrated in Fig.~\ref{fig1b}, these methods handle temporal and inter-variate relationships through separate modules: \textbf{1)} Sequential approach~\cite{wang2024timexer,liu2024itransformer} alternates between modeling temporal and variate dependencies consecutively, where the outcome of one step influences the next. \textbf{2)} Parallel approach~\cite{yu2023dsformer,zhang2023crossformer} independently conducts each modeling process without intermediate interactions and integrates the results only in the final stage.
However, both approaches face significant limitations in explicitly modeling asynchronous interactions, which refer to interactions across different temporal and variate axes, such as in lead-lag relationships.
To overcome this limitation, developing a unified framework that captures temporal and inter-variate dependencies within a single module is essential.

The most straightforward way to process temporal and inter-variate dependencies within a single module is to use the full attention mechanism of the vanilla Transformer~\cite{vaswani2017attention}, as shown in Fig.~\ref{fig1a}.
This approach risks incorporating unnecessary noise, which significantly degrades prediction performance~\cite{leviathan2024selective}.
Based on this observation, we raise the following question: \textit{\textbf{How can we reduce unnecessary noise while simultaneously processing temporal and inter-variate dependencies within a single integrated module?}} 

To address this question, we propose a novel model, Time-Variate Transformer (TiVaT), which concurrently processes temporal and inter-variate dependencies through a single, unified module—the proposed Joint-Axis (JA) attention module.
This module is inspired by deformable attention~\cite{zhu2021deformable}, a method originally introduced in computer vision.
Unlike deformable attention, which focuses on point-level correlations by using offsets to sample key points, our JA attention module shifts the focus to pattern-level sampling.
This module prioritizes patterns such as the temporal flow of key variates and the inter-variate relationships within segments over individual points to construct a candidate pool.
Accordingly, it effectively captures pattern-level information related to the reference points from the perspective of specific timestamps or key variate patterns.
However, some individual data points within the candidate pool may act as noise.

To mitigate this, the JA attention incorporates a novel Distance-aware Time-Variate (DTV) sampling mechanism, which treats all data along the sampled line as candidates rather than directly using them.
DTV sampling projects the candidate and reference points into a 2D embedding space and then extracts the most relevant information based on their distance to the reference point.
This process effectively removes unnecessary noise, improving prediction performance.
Additionally, since this method operates within a visually interpretable 2D space, it enhances the model's explainability.

TiVaT is the first Transformer-based model designed to simultaneously process temporal and inter-variate dependencies through a single unified module, namely the JA attention module.
As illustrated in Fig.~\ref{fig1c}, the module captures asynchronous interactions and cross-variate relationships by focusing on specific timestamps and variates. 
This model demonstrates competitive performance with previous state-of-the-art (SOTA) MTS models, even in complex scenarios where asynchronous interactions and cross-variate relationships are critical. 
The main contributions of this work are as follows:
\begin{itemize}
    \item We present a novel framework, TiVaT, which includes the JA attention module—the first unified mechanism capable of simultaneously processing temporal and variate dependencies.
    \item We propose a novel DTV sampling method that effectively extracts critical patterns based on the learned 2D distance while reducing noise.
    \item TiVaT demonstrates competitive performance against previous SOTA models across a variety of MTS datasets, highlighting its suitability for complex forecasting tasks.
\end{itemize}

%% file: 2_Related_Works.tex
\section{Related Works}
\label{rel}

\begin{figure*}[t]
\begin{center}
\includegraphics[width=\textwidth]{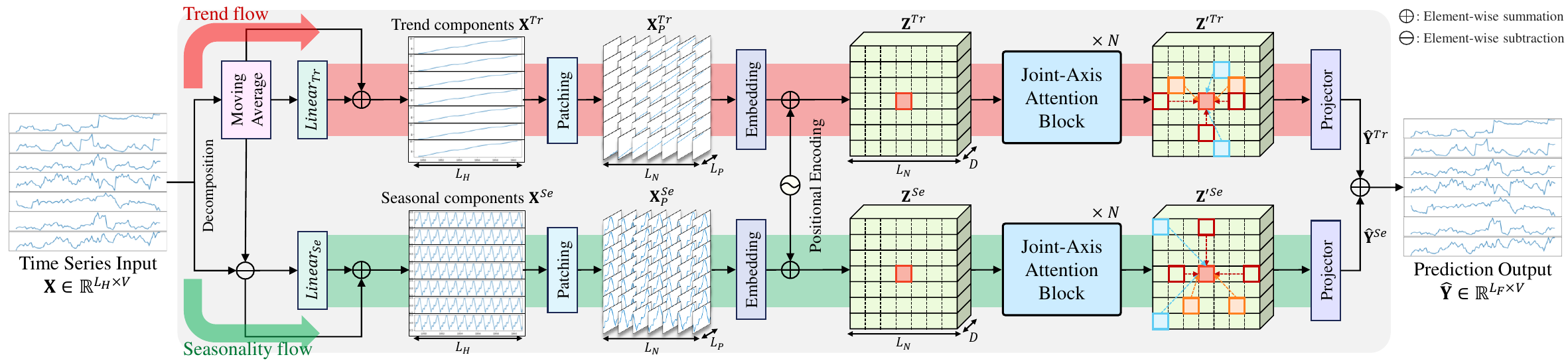}
\end{center}
\caption{Overview of TiVaT.}
\label{fig:overview}
\end{figure*}

Moving beyond traditional approaches, RNNs~\cite{cho2014learning,du2015hierarchical} and CNNs~\cite{bai2018empirical,ismail2020inceptiontime} have demonstrated effectiveness in capturing temporal patterns but struggle with modeling long-term dependencies, often due to architectural limitations and constrained receptive field sizes.
In recent years, Transformer-based models, such as Informer~\cite{zhou2021informer}, Autoformer~\cite{wu2021autoformer}, Non-stationary Transformer~\cite{liu2022non}, and FEDformer~\cite{zhou2022fedformer}, have been adapted as practical tools for temporal modeling in time series data.

MTS forecasting methods can be broadly categorized into CI and CD approaches. CI models, such as DLinear~\cite{zeng2023transformers}, PatchTST~\cite{nie2023a} and TimeMixer~\cite{wang2024timemixer}, treat each variate independently to mitigate overfitting and noise.
However, their inability to explicitly capture interactions between variates limits their effectiveness in datasets characterized by strong inter-variate relationships~\cite{han2024capacity}. To address these limitations, CD models~\cite{zhang2023crossformer,yu2023dsformer,liu2024itransformer,yang2024vcformer,wang2024timexer}, predominantly based on Transformer architectures, employ inter-variate attention mechanisms to effectively model relationships between variates.

These Transformer-based CD models generally adopt one of two strategies: Sequential or Parallel processing, which handle temporal and variate dependencies separately.
Sequential approaches~\cite{wang2024timexer,liu2024itransformer} alternate between modeling temporal and variate dependencies, with the output of one step directly influencing the next. 
For example, iTransformer~\cite{liu2024itransformer} first models temporal dependencies before addressing inter-variate relationships, while TimeXer~\cite{wang2024timexer} focuses on dynamic variate importance through iterative processing. 
Unlike Sequential approaches, Parallel approaches~\cite{yu2023dsformer,zhang2023crossformer} independently process temporal and variate dimensions without intermediate interaction, combining their outputs only at the final stage.
Crossformer~\cite{zhang2023crossformer} and DSformer~\cite{yu2023dsformer} exemplify this category, achieving computational efficiency by decoupling the modeling processes for temporal and variate dimensions.

However, both Sequential and Parallel approaches share a critical limitation: they cannot simultaneously integrate temporal and variate dependencies while effectively capturing asynchronous interactions.
Sequential methods rely on stepwise integration, which isolates cross-axis dependencies and prevents concurrent modeling.
While computationally efficient, parallel methods decouple temporal and variate modeling processes, resulting in fragmented representations that fail to capture holistic cross-axis dynamics.
These common shortcomings highlight the need for a single unified module that can concurrently integrate temporal and variate dependencies to robustly model complex patterns in MTS.

Unlike existing Transformer-based CD models, TiVaT pioneers a JA attention module that simultaneously attends to temporal and variate dimensions in a single unified module. 
By jointly integrating these dependencies, TiVaT effectively addresses the limitations inherent in fragmented methods.
This novel mechanism enables the model to capture complex inter-variate dependencies, such as lead-lag relationships, which were previously challenging to model.

%% file: 3_Methodology.tex
\section{Methodology}
\label{method}

\begin{figure*}[h]
\begin{center}
\includegraphics[width=\textwidth]{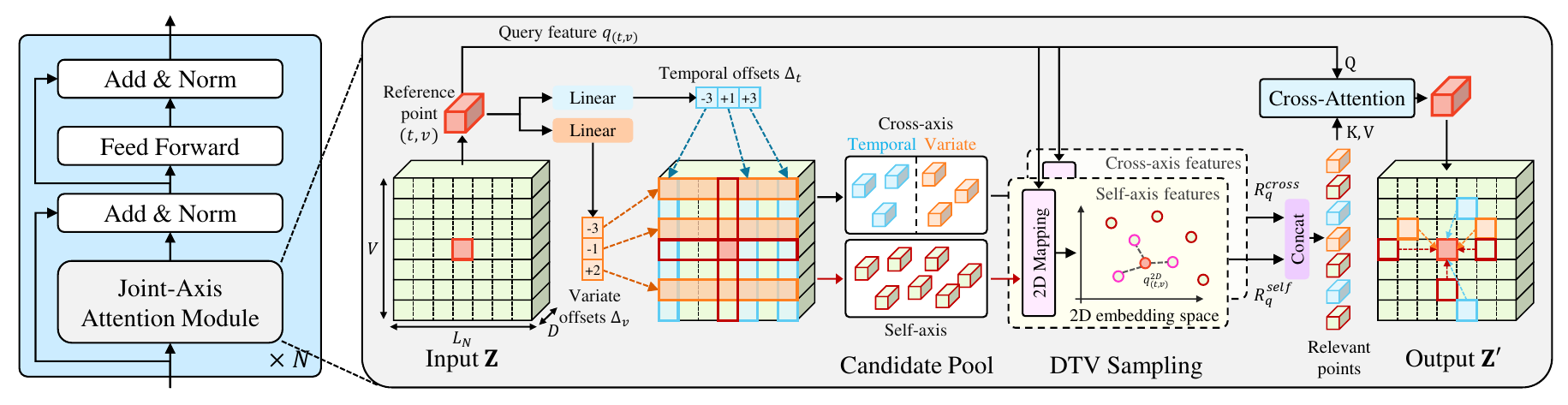}
\end{center}
\caption{Joint-Axis Attention Block.}
\label{fig:jablock}
\end{figure*}

\subsection{Backgrounds}

\paragraph{Problem Definition.}
MTS forecasting is a task that leverages historical data to predict future values for each variate.
Formally, given historical data $X=\{x_{T-L_H+1}, ..., x_{T} \} \in \mathbb{R}^{L_H \times V}$, where $V$ is the number of variates and $L_H$ is the number of time steps up to a given time point $T$, the objective is to predict $L_F$ time steps of future data $Y=\{x_{T+1}, ..., x_{T+L_F} \} \in \mathbb{R}^{L_F \times V}$.
For a data point $X_{(t,v)}$, we define a dependency with another point $X_{(t',v')}$ as $\mathcal{D}_{(t,v)\leftarrow(t',v')}$, where the arrow indicates the direction of dependency from $(t', v')$ to $(t, v)$.
Note that we denote temporal data points for a single variate $v$ as $X_{(:,v)} \in \mathbb{R}^{L_H}$ and variate data points at a specific time step $t$ as $X_{(t,:)} \in \mathbb{R}^V$.

\paragraph{Motivation Formulation.}
Existing Transformer-based approaches~\cite{zhang2023crossformer,yang2024vcformer,wang2024timexer,liu2024itransformer,yu2023dsformer} focus on either temporal dependencies $\mathcal{D}_{(t,v)\leftarrow(t',v')}$, where $t' \neq t$ and $v' = v$, or inter-variate dependencies, where $t' = t$ and $v' \neq v$, for a data point $X_{(t,v)}$. These methods treat temporal and variate relationships separately.
Consequently, they struggle to capture the intricate patterns in asynchronous dependencies $\mathcal{D}_{(t,v)\leftarrow(t',v')}$ for $X_{(t,v)}$, where $t' \neq t$ and $v' \neq v$, including lead-lag relationships where $t' < t$ and $v' \neq v$.
Our TiVaT is motivated by these limitations.

\subsection{Architecture Overview}
Fig.~\ref{fig:overview} describes an overview of TiVaT, designed to effectively capture intricate and asynchronous cross-axis interactions across both variate and temporal axes simultaneously through the JA attention blocks.
First, TiVaT applies the seasonal-trend decomposition method to the normalized MTS data to reduce its complexity.
Following previous works~\cite{cleveland1990stl,wang2024timemixer}, the input sequence for each variate is decomposed into two components: the moving average, which represents the trend $X^{Tr} \in \mathbb{R}^{L_H \times V}$, and the remainder, which is treated as seasonality $X^{Se} \in \mathbb{R}^{L_H \times V}$.
In addition, to preserve the temporal characteristics and enhance the representation of their patterns, each component is processed through individual linear layers $Linear_i(\cdot)$, where $i\in\{Tr, Se\}$, with residual connections, as follows:
\begin{equation}
    \begin{split}
        X^{Tr} &= MA(X),  \\
        X^{Se} &= X - X^{Tr},     \\
        \hat{X}^{Tr} &= X^{Tr} + Linear_{Tr}(X^{Tr}),  \\
        \hat{X}^{Se} &= X^{Se} + Linear_{Se}(X^{Se}),
    \end{split}
\end{equation}
where $MA$ represents the moving average for the temporal axis for each variate.
% As shown in Fig.~\ref{fig:overview}, the decomposed components $X^{Tr}$ and $X^{Se}$ are individually processed through sibling architectures to reduce confusion arising from the difference of long-term and short-term properties.
The decomposed components $\hat{X}^{Tr}$ and $\hat{X}^{Se}$ are individually processed through sibling architectures to reduce confusion arising from the difference of long-term and short-term properties.

Each architecture consist of an embedding layer, $N$ JA attention blocks, and a projection layer.
For the embedding layer, we adopt the patch embedding method~\cite{nie2023a} to alleviate long-term dependencies and enhance local temporal information.
When each component is divided into patches of length $L_P$ and a stride of $S$ along the temporal axis, the input length $L_H$ is reduced to $L_N=\lfloor{\frac{L_H-L_P}{S}}\rfloor+2$ and a new dimension corresponding to the patch length $L_P$ is introduced, resulting in the patched input $X_P\in \mathbb{R}^{L_N \times V \times L_P}$.
Subsequently, Input tokens $Z\in \mathbb{R}^{L_N \times V \times D}$ are generated by feeding the patched input into a linear layer and adding the positional encoding.

TiVaT learns the complex cross-axis relationships in the MTS data based on the input tokens $Z$ using the JA attention blocks.
The intermediate predictions $\hat{Y}^{Tr}$ and $\hat{Y}^{Se}$ for trend and seasonality components are generated through a linear layer-based projector $Proj:\mathbb{R}^{L_N \times V \times D}\rightarrow\mathbb{R}^{L_F \times V}$.
The final prediction $\hat{Y}\in \mathbb{R}^{L_F \times V}$ is obtained by aggregating these intermediate predictions through an element-wise sum $\oplus$, as follows:
\begin{equation}
\begin{split}
    % \hat{Y}^{Tr} &= Proj\bigl(Enc\bigl(Emb(X^{Tr}) + PE\bigr)\bigr), \\
    % \hat{Y}^{Se} &= Proj\bigl(Enc\bigl(Emb(X^{Se}) + PE\bigr)\bigr), \\
    \hat{Y}^{j} &= Proj^{j}\bigl(Enc^{j}\bigl(Emb^{j}(\hat{X}^{j}) + PE\bigr)\bigr) \\
    \hat{Y} &= \hat{Y}^{Tr} \oplus \hat{Y}^{Se},
\end{split}
\label{eq:final_prediction}
\end{equation}
where $Enc^{j}$ and $Emb^{j}$ represent the JA attention blocks and patch embedding for $j\in\{Tr, Se\}$, respectively, and $PE$ is the positional encoding.

\subsection{Joint-Axis Attention Block}
\label{sec:ja_attention}
As shown in Fig.~\ref{fig:jablock}, the JA attention block adopts a Transformer encoder block structure~\cite{vaswani2017attention}, replacing the standard self-attention mechanism with the JA attention module.
The JA attention module is inspired by the offset mechanism of deformable attention~\cite{zhu2021deformable}, enabling it to capture complicated relationships including asynchronous dependencies $\mathcal{D}_{(t,v)\leftarrow(t',v')}$.
This makes the JA attention a single unified module capable of simultaneously processing temporal and variate dependencies.

For a query feature $q_{(t,v)}\in\mathbb{R}^{D}$ at each reference point $(t,v)$ on the feature map $Z$, our module extracts offsets using linear layers.
These offsets represent the displacement from the reference point along both the time and variate axes, allowing for the simultaneous consideration of interactions across these dimensions.
The original deformable attention is a point-based method that considers the locality of images—the correlation between neighboring pixels—and samples only key points related to the reference point using offsets.
However, in time series data, identifying meaningful patterns requires focusing on specific timestamps or variates rather than spatial locality~\cite{zhou2015feature,pan2015feature}. 
Therefore, the JA attention extends the concept of offsets to define them as guidelines along temporal and variate axes. 

In other words, the JA attention module uses offsets to construct candidate pools for sampling features relevant to the query.
However, these candidate pools may still contain irrelevant noise concerning the query.
To address this issue, we propose a novel top-$K$ sampling method, DTV sampling, which filters out features that contribute to the query representation based on the Euclidean distance in the 2D embedding space, thereby minimizing noise.
Finally, all sampled relevant features are integrated and used to update the corresponding query feature through a cross-attention layer.
The JA attention module refines the feature map $Z$ by replacing the original query features with the updated ones at their respective locations, enhancing the representation capacity of the feature map.

\paragraph{Configuration of Candidate Pools.}

As illustrated in Fig.~\ref{fig:jablock}, we construct two types of candidate pools for the query feature $q_{(t,v)}$ at each reference point $(t,v)$: 
\emph{(i) the self-axis pool} and \emph{(ii) the cross-axis pool}.
The self-axis pool covers features $Z_{(t',v')}$ at $t'=t$ or $v'=v$ for the query feature $q_{(t,v)}$, and the cross-axis pool consists of features at $t'\neq t$ and $v'\neq v$.
% In MTS analysis, the value of a reference point $(t,v)$ is often considered to be most related to the values of all time steps of the same variate ($Z_{(:,v)}$) and the values of all variates at the same time step ($Z_{(t,:)}$).
% In MTS analysis, the value of a reference point $(t,v)$ is often considered to be most related to the historical values of its own variate ($Z_{(:,v)}$) and the values of other variates at the same time step ($Z_{(t,:)}$)~\cite{hochreiter1997long,tealab2018time}.
In MTS analysis, the value at a reference point $(t,v)$ is often considered to be most relevant to the historical values of its own variate ($Z_{(:,v)}$) and the values of other variates at the same time step ($Z_{(t,:)}$)~\cite{hochreiter1997long,tealab2018time}.
% Thus, we additionally construct the self-axis pool to incorporate this inductive bias during training our model.
Thus, we construct the self-axis pool to incorporate this inductive bias.
For the cross-axis pool, the temporal and variate offsets, $\Delta_{t}$ and $\Delta_{v}$, are extracted by passing the query feature through their respective linear layers.
Initially, $\Delta_{t}$ and $\Delta_{v}$ are determined as unconstrained real numbers and then normalized into their respective temporal and variate ranges.
This process ensures that the offsets cover the entire area of the feature map $Z$.
These offsets serve as guidelines to construct the cross-axis pool.
When a temporal offset $\Delta_{t}$ is determined for the query feature $q_{(t,v)}$, all variate feature vectors $Z_{(t+\Delta_{t},:)}$ at the time step $t+\Delta_{t}$ are included in the cross-axis pool.
Similarly, when a variate offset $\Delta_{v}$ is obtained, all temporal feature vectors $Z_{(:,v+\Delta_{v})}$ for the variate $v+\Delta_{v}$ are added to the cross-axis pool.
As relevant information differs depending on the number of variates $V$ or patch length $L_P$, we determine the number of $\Delta_{t}$ and $\Delta_{v}$ by hyperparameters $p_t$ and $p_v$, which represent the proportions of the number of elements on each axis, respectively.

\paragraph{Distance-aware Time-Variate Sampling.}

To mitigate the reflection of irrelevant noise from candidate pools into query features, we propose the DTV sampling method based on top-$K$ sampling. 
DTV sampling uses Euclidean distances in the 2D embedding space as a criterion to select $K$ features most closely related to the queries from the candidate pools.
This sampling method operates in a visible embedding space, enhancing both the model's interpretability and sampling effectiveness.
DTV sampling is applied separately to the self-axis and cross-axis pools, and this strategy was determined based on our experiments in Supp.~\ref{sec:supp_sep}.
For each pool, DTV sampling first projects the query feature $q_{(t,v)}$ and features $Z_{(t',v')}$ in the pool into a 2D embedding space, resulting in $q^\text{2D}_{(t,v)}$ and $Z^\text{2D}_{(t',v')}$, respectively.
Subsequently, the indices $I_q$ of the relevant points in the pool are determined based on the Euclidean distance, denoted as $Dist$, as follows:
\begin{equation}
    \label{eq:sampling}
    I_q = \text{argtop}K_{(t',v')}\bigl({Dist}(q_{(t,v)}^{\text{2D}},\,Z_{(t',v')}^{\text{2D}})\bigr),
\end{equation}
where argtop$K$ represents a function that extracts $K$ indices $(t',v')$ corresponding to the shortest distance from their query.
The relevant feature vectors $R_q\in\mathbb{R}^{K \times D}$ of the pool are sampled at the $I_q$ positions on the feature map.
% We denote the relevant features from the self- and cross-axis pools as $R_q^{self}$ and $R_q^{cross}$, respectively.
The relevant features from the self-axis and cross-axis pools are denoted as $R_q^{self}$ and $R_q^{cross}$, respectively.
All features in $R_q^{self}$ and $R_q^{cross}$ are reflected in their query feature by using them as key and value features in the cross-attention.

\input{table/avg_result_table}  
% \paragraph{Attention Operation.}
% \subsubsection{Feature Map Refinement}
\paragraph{Query-level Cross Attention.}
% Finally, the sampled features $R_q^{self}$ and $R_q^{cross}$ are integrated and injected into the query feature $q_{(t,v)}$, updating it to reflect relationships with other points.
% This process generates a new feature map $Z'$ composed of updated queries $q'_{(t,v)}$ for all $(t,v)$, capable of representing complex interactions in MTS data.
Finally, the sampled features $R_q^{self}$ and $R_q^{cross}$ are integrated and injected into the query feature $q_{(t,v)}$ to update it, reflecting relationships with other points. This process generates a new feature map $Z'$ , composed of updated queries $q'_{(t,v)}$ for all $(t,v)$, which represents complex interactions in MTS data.
% Unlike traditional attention-based models, this approach effectively captures relationships that include information across different time points and variates, such as lagged points $Z_{(t',v')}$, which exist at $t' < t$ and $v' \neq v$.
This approach effectively captures relationships that include information across different time points and variates, such as lagged points $Z_{(t',v')}$, which exist at $t' < t$ and $v' \neq v$.
For all reference points $(t,v)$, we inject the sampled feature vectors into the query using a cross-attention layer.
In the cross-attention layer, $q_{(t,v)}$ serves as the query, while the selected feature vectors $R_q^{self}$ and $R_q^{cross}$ are concatenated and used as the key and value.
The query $\mathbf{Q}$, key $\mathbf{K}$, and value $\mathbf{V}$ are generated through linear projections, as follows:
\begin{align}
\begin{split}
   % \mathbf{Q} &= Proj^q\bigl(q_{(t,v)}\bigr) \in \mathbb{R}^{1 \times D},\\
   % \mathbf{K} &= Proj^k\bigl([\,R_q^{self} \,\|\, R_q^{cross}]\bigr) \in \mathbb{R}^{2K \times D},\\
   % \mathbf{V} &= Proj^v\bigl([\,R_q^{self} \,\|\, R_q^{cross}]\bigr) \in \mathbb{R}^{2K \times D},
   \mathbf{Q} &= Proj^q\bigl(q_{(t,v)}\bigr),\\
   \mathbf{K} &= Proj^k\bigl([\,R_q^{self} \,\|\, R_q^{cross}]\bigr),\\
   \mathbf{V} &= Proj^v\bigl([\,R_q^{self} \,\|\, R_q^{cross}]\bigr),
\end{split}
\end{align}
where $[\cdot \,\|\, \cdot]$ indicates concatenation and $Proj^i$ (for $i \in {q,k,v}$) refer to separate linear layers for the query, key, and value, respectively.
From these operations, as shown in Eq.~\ref{eq:crossattn}, the updated query feature $q'_{(t,v)}$ is extracted based on attention scores, which are computed using the scaled dot product. 
\begin{equation}
    \label{eq:crossattn}
    q'_{(t,v)} = \mathrm{Softmax}\Bigl(\mathbf{Q} \cdot \mathbf{K}^\mathsf{T} / {\sqrt{D}}\Bigr) \cdot \mathbf{V},
\end{equation}
where ( $\cdot$ ) represents the dot product.
By integrating the sampled features into the query, the JA attention mechanism enhances the feature map’s ability to represent both temporal and variate interactions.

%% file: table/avg_result_table.tex
\renewcommand{\arraystretch}{} 
\begin{table*}[t!]
\large  
\resizebox{\textwidth}{!}{%
\begin{tabular}{c|cccccccccccccccccccccccc}
\toprule
                 
\multicolumn{1}{c|}{\multirow{2}{*}{Models}}   & \multicolumn{2}{c|}{\textbf{TiVaT}} & \multicolumn{2}{c|}{TimeXer}       & \multicolumn{2}{c|}{VCformer}                              & \multicolumn{2}{c|}{iTransformer}  & \multicolumn{2}{c|}{TimeMixer} & \multicolumn{2}{c|}{DSformer}    & \multicolumn{2}{c|}{PatchTST}   & \multicolumn{2}{c|}{Crossformer}     & \multicolumn{2}{c|}{TimesNet}      & \multicolumn{2}{c|}{DLinear} & \multicolumn{2}{c|}{FEDformer}  & \multicolumn{2}{c}{Autoformer}  \\

\multicolumn{1}{c|}{}         & \multicolumn{2}{c|}{\textbf{(Ours)}}  & \multicolumn{2}{c|}{(2024)}          & \multicolumn{2}{c|}{(2024)}                                 & \multicolumn{2}{c|}{(2024)}          & \multicolumn{2}{c|}{(2024)}          & \multicolumn{2}{c|}{(2023)}          & \multicolumn{2}{c|}{(2023)}& \multicolumn{2}{c|}{(2023)} & \multicolumn{2}{c|}{(2023)}      & \multicolumn{2}{c|}{(2023)}      & \multicolumn{2}{c|}{(2022)}          & \multicolumn{2}{c}{(2021)}    \\ \midrule

\multicolumn{1}{c|}{Metrics}   & MSE    & \multicolumn{1}{c|}{MAE}   & MSE    & \multicolumn{1}{c|}{MAE} & MSE    & \multicolumn{1}{c|}{MAE} & MSE    & \multicolumn{1}{c|}{MAE} & MSE   & \multicolumn{1}{c|}{MAE}   & MSE   & \multicolumn{1}{c|}{MAE}                          & MSE   & \multicolumn{1}{c|}{MAE}   & MSE   & \multicolumn{1}{c|}{MAE}   & MSE   & \multicolumn{1}{c|}{MAE}   & MSE   & \multicolumn{1}{c|}{MAE}   & MSE   & \multicolumn{1}{c|}{MAE}   & MSE          & MAE          \\ \midrule

\multicolumn{1}{c|}{ETTh1 (7)}    & \textcolor{red}{\textbf{0.434}}  & \multicolumn{1}{c|}{\textcolor{red}{\textbf{0.435}}} & \textcolor{blue}{0.437} & \multicolumn{1}{c|}{\textcolor{blue}{0.437}} & 0.439 & \multicolumn{1}{c|}{\textcolor{blue}{0.437}}                        & 0.454 & \multicolumn{1}{c|}{0.447} & 0.447 & \multicolumn{1}{c|}{0.440} & \textcolor{blue}{0.437}& \multicolumn{1}{c|}{0.441} & 0.469 & \multicolumn{1}{c|}{0.454}& 0.529 & \multicolumn{1}{c|}{0.522} &    0.458     &  \multicolumn{1}{c|}{0.450}&0.456&\multicolumn{1}{c|}{0.452}   & 0.440 & \multicolumn{1}{c|}{0.460} & 0.496& 0.487     \\

\multicolumn{1}{c|}{ETTh2 (7)}    & 0.370  & \multicolumn{1}{c|}{0.400} & \textcolor{blue}{0.367} & \multicolumn{1}{c|}{\textcolor{blue}{0.396}} & 0.377 & \multicolumn{1}{c|}{0.403}                        & 0.383 & \multicolumn{1}{c|}{0.407} & \textcolor{red}{\textbf{0.364}} & \multicolumn{1}{c|}{\textcolor{red}{\textbf{0.395}}}  & 0.396 & \multicolumn{1}{c|}{0.418} & 0.387 & \multicolumn{1}{c|}{0.407} & 0.942 & \multicolumn{1}{c|}{0.684}& 0.414        & \multicolumn{1}{c|}{0.427}&0.559&\multicolumn{1}{c|}{0.515}  & 0.437 & \multicolumn{1}{c|}{0.449} & 0.450&0.459          \\

\multicolumn{1}{c|}{ETTm1 (7)}    & \textcolor{red}{\textbf{0.380}}  & \multicolumn{1}{c|}{\textcolor{blue}{0.397}} & 0.382 & \multicolumn{1}{c|}{\textcolor{blue}{0.397}} & 0.387 & \multicolumn{1}{c|}{\textcolor{blue}{0.397}}                        & 0.407 & \multicolumn{1}{c|}{0.410} & \textcolor{blue}{0.381} & \multicolumn{1}{c|}{\textcolor{red}{\textbf{0.395}}} & 0.389 & \multicolumn{1}{c|}{0.401} & 0.387 & \multicolumn{1}{c|}{0.400}& 0.513 & \multicolumn{1}{c|}{0.496}  & 0.400       & \multicolumn{1}{c|}{0.406}&0.403&\multicolumn{1}{c|}{0.407}    & 0.448 & \multicolumn{1}{c|}{0.452} & 0.588 & 0.517          \\

\multicolumn{1}{c|}{ETTm2 (7)}    & 0.276  & \multicolumn{1}{c|}{0.325} & \textcolor{red}{\textbf{0.274}} & \multicolumn{1}{c|}{\textcolor{red}{\textbf{0.322}}} & 0.285 & \multicolumn{1}{c|}{0.330}                        & 0.288 & \multicolumn{1}{c|}{0.332} & \textcolor{blue}{0.275} & \multicolumn{1}{c|}{\textcolor{blue}{0.323}} & 0.312 & \multicolumn{1}{c|}{0.351} & 0.281 & \multicolumn{1}{c|}{0.326} & 0.757 & \multicolumn{1}{c|}{0.610} & 0.291        &  \multicolumn{1}{c|}{0.333}&0.350&\multicolumn{1}{c|}{0.401}    & 0.305 & \multicolumn{1}{c|}{0.349} & 0.327& 0.371       \\

\multicolumn{1}{c|}{Exchange (8)} & \textcolor{red}{\textbf{0.349}}  & \multicolumn{1}{c|}{\textcolor{red}{\textbf{0.398}}} & 0.422 & \multicolumn{1}{c|}{0.416} & 0.355 & \multicolumn{1}{c|}{\textcolor{blue}{0.402}} & 0.360 & \multicolumn{1}{c|}{0.403} & 0.397 & \multicolumn{1}{c|}{0.414} & 0.394 & \multicolumn{1}{c|}{0.425} & 0.367 & \multicolumn{1}{c|}{0.404}& 0.940 & \multicolumn{1}{c|}{0.707} & 0.416  & \multicolumn{1}{c|}{0.443}  & \textcolor{blue}{0.353}  & \multicolumn{1}{c|}{0.414}  & 0.519 & \multicolumn{1}{c|}{0.429} &0.613 &0.539    \\

\multicolumn{1}{c|}{Weather (21)}  & \textcolor{red}{\textbf{0.240}}  & \multicolumn{1}{c|}{\textcolor{red}{\textbf{0.270}}} & \textcolor{blue}{0.241} & \multicolumn{1}{c|}{\textcolor{blue}{0.271}} & 0.258 & \multicolumn{1}{c|}{0.282}                         & 0.258 & \multicolumn{1}{c|}{0.278} & \textcolor{red}{\textbf{0.240}} & \multicolumn{1}{c|}{\textcolor{blue}{0.271}}& 0.276 & \multicolumn{1}{c|}{0.304} & 0.259 & \multicolumn{1}{c|}{0.281}  & 0.259  & \multicolumn{1}{c|}{0.315} & 0.259  & \multicolumn{1}{c|}{0.287} & 0.265  & \multicolumn{1}{c|}{0.317}  & 0.309 & \multicolumn{1}{c|}{0.360} & 0.338&0.382       \\

\multicolumn{1}{c|}{ECL (321)}      & \textcolor{red}{\textbf{0.166}}  & \multicolumn{1}{c|}{\textcolor{red}{\textbf{0.262}}} &\textcolor{blue}{0.171} & \multicolumn{1}{c|}{0.270} & 0.180 & \multicolumn{1}{c|}{\textcolor{blue}{0.267}} & 0.178 & \multicolumn{1}{c|}{0.270} & 0.182 & \multicolumn{1}{c|}{0.272} & 0.196& \multicolumn{1}{c|}{0.289} & 0.205 & \multicolumn{1}{c|}{0.290}  & 0.244  & \multicolumn{1}{c|}{0.334} & 0.192 & \multicolumn{1}{c|}{0.295} & 0.212  & \multicolumn{1}{c|}{0.300}   & 0.214 & \multicolumn{1}{c|}{0.327} & 0.227 &0.338        \\

\multicolumn{1}{c|}{Traffic (862)}  & \textcolor{blue}{0.437}  & \multicolumn{1}{c|}{0.297} & 0.466 & \multicolumn{1}{c|}{\textcolor{blue}{0.287}} & 0.550 & \multicolumn{1}{c|}{0.304}                         & \textcolor{red}{\textbf{0.428}} & \multicolumn{1}{c|}{\textcolor{red}{\textbf{0.282}}} & 0.484 & \multicolumn{1}{c|}{0.297} & 0.563 & \multicolumn{1}{c|}{0.355} & 0.481 & \multicolumn{1}{c|}{0.304}  & 0.550  & \multicolumn{1}{c|}{0.304} & 0.620  & \multicolumn{1}{c|}{0.336} &  0.625 & \multicolumn{1}{c|}{0.383}   & 0.610 & \multicolumn{1}{c|}{0.376} & 0.628&0.379   \\ \bottomrule
\end{tabular}%
}
% \caption{Multivariate long-term time series forecasting results, with the value in parentheses indicating the number of variates in each dataset.}
\caption{Multivariate long-term time series forecasting results, with the number of variates in each dataset indicated in parentheses.}
\label{tbl:main_results}
\end{table*}

%% file: 4_Experiments.tex
\section{Experiments}
\label{exp}
% \paragraph{Dataset.} 
\subsection{Experimental Settings}
\paragraph{Dataset and Metrics.} 
Our experimental evaluation utilizes eight real-world datasets that are widely used in time-series forecasting research, ensuring a rigorous and comprehensive comparison with SOTA models.
These datasets include ECL, ETT (with four subsets), Exchange, Traffic, and Weather, following Autoformer~\cite{wu2021autoformer} for long-term forecasting.
For the ablation study, experiments are conducted on the ETTh1 (Electricity), Exchange (Economy), and Weather(Weather) datasets to analyze the effectiveness of the proposed model across various domains.
% For the ablation study, experiments were conducted on the ETTh1(Electricity), Exchange(Economy), and Weather(Weather) datasets to analyze the effectiveness of the proposed model across various domains.
Detailed configurations for each dataset are provided in the Supp.~\ref{sec:supp_data}.
In this paper, we evaluate all models using mean squared error (MSE) and mean absolute error (MAE), consistent with prior works.

\paragraph{Baselines.}
% We use the following popular models for MTS forecasting as baselines:
We select 11 well-acknowledged MTS forecasting models as baselines, including:  
TimeXer~\cite{wang2024timexer}, VCformer~\cite{yang2024vcformer}, iTransformer~\cite{liu2024itransformer}, TimeMixer~\cite{wang2024timemixer}, DSformer~\cite{yu2023dsformer}, PatchTST~\cite{nie2023a}, Crossformer~\cite{zhang2023crossformer}, TimesNet~\cite{wu2023timesnet}, Dlinear~\cite{zeng2023transformers}, FEDformer~\cite{zhou2022fedformer}, and Autoformer~\cite{wu2021autoformer}.

% We carefully select eleven previously successful forecasting models as our benchmarks, categorized into two groups: 
% We carefully select eight well-known forecasting models as our benchmarks, which can be categorized into two groups based on their methodologies:
% (1) CD models: DeformableTST~\cite{luodeformabletst}, TimeXer~\cite{wang2024timexer}, VCformer~\cite{yang2024vcformer}, iTransformer~\cite{liu2023itransformer}, Crossformer~\cite{zhang2023crossformer}, TimesNet~\cite{wu2023timesnet}.
% (2) CI models: PatchTST~\cite{nie2023a}, TimeMixer~\cite{wang2024timemixer}, Dlinear~\cite{zeng2023transformers}.

% \paragraph{Unified experiment settings.} 
% For a fair comparison, we compare our model's performance with the results reported in previous studies. In cases where the input sequence length differs, we conduct experiments using the parameters provided by the respective papers to ensure consistency in performance comparison. Furthermore, to evaluate the generalization ability of the model, all experimental results are reported as the average performance across forecasting lengths of 96, 192, 336, and 720.

% \paragraph{Implementation Details.} 
% All experiments are conducted using PyTorch~\cite{paszke2017automatic} on multiple NVIDIA A100 GPUs with 80GB of memory. The model training is performed using MSE loss. 
% Additionally, the number of reference points is determined by $p_{t}$, which selects a percentage of the time axis, and $p_{v}$, which selects a percentage of the variate axis, ensuring both efficiency and optimal performance depending on the dataset.
% % $K$
\paragraph{Implementation Details.}
% To ensure fairness in performance comparison, we compare our model's results with those reported in baseline studies.
For fair performance comparison, we compare our model's results with those reported in baseline studies.
This study employs a fixed lookback length $L_H=96$ and evaluate the average performance across prediction lengths $L_F\in\{96, 192, 336, 720\}$ for all experiments.
%, following the settings used in iTransformer~\cite{liu2024itransformer}.
The optimal hyperparameters, such as $p_t$, $p_v$, and $K$, are determined based on the characteristics of each dataset and the target prediction length.
% Detailed implementation settings are provided in the Supplementary Material, and we train the model using the MSE loss function.
% We provide detailed implementation settings in the Supp.~\ref{sec:supp_training} and train the model using the MSE loss function.
We provide detailed implementation settings in the Supp.~\ref{sec:supp_training}.
The model is trained using the MSE loss function.
% All experiments are conducted using PyTorch~\cite{paszke2017automatic} on NVIDIA A100 GPUs (80GB memory) with multiple GPUs utilized for parallel computation.
All experiments are conducted using PyTorch~\cite{paszke2017automatic} on NVIDIA A100 GPUs (80GB memory), leveraging multiple GPUs for parallel computation.

\subsection{Experimental Results}
\label{Results}

Table~\ref{tbl:main_results} presents the long-term forecasting results, where the best and second-best results are highlighted in \textbf{\textcolor{red}{red}} and \textcolor{blue}{blue}, respectively.
A lower MSE/MAE indicates a more accurate prediction. 
TiVaT achieves overall SOTA performance across diverse benchmark datasets, showcasing its versatility and advanced modeling capabilities.

TiVaT demonstrates superior results on ETTh1, ETTm1, and Exchange, where temporal dependencies are critical due to the relatively small number of variates, as described in Table~\ref{tbl:main_results}.
% 1. TiVaT demonstrated superior results on ETTh1, ETTm1, and Exchange datasets, where CI models typically outperform CD models in scenarios with a small number of variates. Contrary to this trend, TiVaT’s exceptional performance highlights the importance of capturing asynchronous dependencies in low-dimensional datasets.
% 2. TiVaT demonstrates superior performance on ETTh1, ETTm1, and Exchange datasets, outperforming not only CD models such as TimeXer and iTransformer but also CI models like TimeMixer and PatchTST. These results underscore the importance of effectively modeling asynchronous dependencies even in low-dimensional datasets.
% It surpasses CD models such as TimeXer and iTransformer and even outperforms CI models like TimeMixer and PatchTST, which are specifically designed to capture temporal dependencies.
% These results highlight TiVaT’s ability to effectively model temporal patterns and deliver highly accurate predictions in low-dimensional datasets.
In particular, our method significantly improves performance for Exchange, which has more complexity due to non-stationary characteristics~\cite{wu2021autoformer}.
This demonstrates that our approach, which can capture asynchronous interactions and reduce noise, is especially effective for handling non-stationary data.
% 
% TiVaT demonstrates outstanding results on ETTh1, ETTm1, and Exchange, where temporal dependencies are emphasized due to the relatively small number of variates.
% It surpasses CD models such as TimeXer~\cite{wang2024timexer} and iTransformer~\cite{liu2024itransformer} and even outperforms CI models like TimeMixer~\cite{wang2024timemixer} and PatchTST~\cite{nie2023a}, which are specifically designed to capture temporal dependencies.
% These results highlight TiVaT’s ability to effectively model temporal patterns and deliver highly accurate predictions in low-dimensional datasets.

On high-dimensional datasets, TiVaT further demonstrates its superiority by achieving SOTA performance on the Weather and ECL benchmarks, outperforming other CD models such as TimeXer, iTransformer, and VCformer, where inter-variate dependencies and complex patterns are critical.
Additionally, TiVaT secures the second-best MSE result on the Traffic dataset, which is characterized by dynamic multivariate relationships and temporal variations.
These results emphasize TiVaT's capability to handle asynchronous dependencies with consistency across varying dataset complexities.
In summary, TiVaT’s strong performance across diverse datasets highlights its effectiveness in capturing temporal and inter-variate dependencies, positioning it as a reliable solution for complex MTS forecasting.

\subsection{Analysis}

\subsubsection{Ablation on Joint-Axis Attention Module} 
% JA Attention is designed to integrate temporal and variate dependencies seamlessly, empowering the model to effectively identify and utilize complex variate-temporal interactions that are often overlooked by traditional methods.
% ------자세한 설명 버전------Method (a) and method (b) were both evaluated using the same hyperparameters as TiVaT to ensure a fair comparison. Method (a) replaces the JA Attention block with the vanilla Transformer's encoder block, where every feature serves as the key and value in the attention mechanism for a given query. Method (b), on the other hand, replaces the JA Attention block with the encoder block provided by Crossformer, which sequentially performs attention operations—first along the temporal axis and then along the variate axis—thereby capturing temporal and inter-variate dependencies.

To validate the effectiveness of JA attention in simultaneously processing temporal and inter-variate dependencies without relying on the full set of features, we conduct a comparative analysis against two alternative methods: (a) Full Attention and (b) Two-Stage Attention. 
In both methods, the overall structure of TiVaT was preserved, except for replacing the JA attention block.
Table~\ref{tbl:ablation_JA_attn} shows the comparison results for methods (a), (b), and the proposed JA attention (c).

% Method (c) consistently outperformed method (a), which performs computations over all features using the vanilla Transformer's encoder block~\cite{vaswani2017attention}, across benchmark datasets.
% Across benchmark datasets, method (c) consistently outperformed method (a), which performs computations over all features using the vanilla Transformer's encoder block~\cite{vaswani2017attention}.
Across benchmark datasets, our method consistently outperforms the full attention, which performs computations over all features using the vanilla Transformer's encoder block~\cite{vaswani2017attention}.
% This demonstrates that JA attention effectively extracts key features from the entire feature map while reducing unnecessary noise, leading to improved performance.
This result indicates that JA attention effectively extracts key features from the entire feature map while reducing unnecessary noise, leading to improved performance.
% This infers that JA attention, which selectively reflects features, effectively reduces  unnecessary features such as noise compared to full attention.
Compared to method (b), which separately models temporal and inter-variate dependencies using Crossformer's encoder block~\cite{zhang2023crossformer}, (c) also achieves improved results across benchmark datasets. 
In particular, on the Weather dataset, characterized by its relatively high variate count, (c) showed the largest performance improvement, outperforming method (b) by 13.04\%.
%of 13.04\% over (b).
%Notably, method (c) outperforms method (b) by 13.04\% on the Weather dataset, which is characterized by its relatively high number of variates.
These findings highlight the critical role of simultaneously modeling temporal and inter-variate dependencies with a single unified module instead of handling them separately to capture asynchronous interactions effectively.

\input{table/ablation_JA_attention}

\subsubsection{Ablation on DTV Sampling}

% To evaluate the impact of sampling strategies on feature selection, we compare two approaches: (a) random sampling and (b) the proposed DTV sampling.
To evaluate the impact of sampling strategies on feature selection, we compare proposed DTV sampling (c) with two approaches: (a) without sampling and (b) random sampling.
The experimental results presented in Table~\ref{tbl:ablation_DTV} demonstrate that DTV sampling consistently outperforms both (a) and (b) across multiple datasets.
These findings highlight the effectiveness of DTV sampling in identifying and extracting semantically relevant features from candidate pools, which ultimately enhances model performance.

Further analysis comparing (a) and (c) reveals that DTV sampling performs better on benchmark datasets.
% This indicates that combining the guidelines with DTV sampling is more effective than relying solely on the guidelines.
This indicates that combining guidelines with DTV sampling is more effective than using the guidelines alone.
Furthermore, this suggests that specific individual data points in the candidate pools introduce noise, which the DTV sampling method effectively mitigates, thereby improving overall performance.

Additionally, comparing methods (b) and (c) reinforces the importance of DTV sampling, showing that it consistently outperforms random sampling.
% In addition, comparing (b) and (c) reinforces the importance of DTV sampling. The results show that DTV sampling outperforms random sampling.
Unlike random sampling, which selects features without considering their relevance, DTV sampling aligns features with reference points through a learned 2D embedding space.
This approach ensures that critical patterns are identified and irrelevant information is excluded, further validating the robustness of the proposed method.
% Therefore, DTV sampling is pivotal for effective feature selection, directly improving the model’s performance.
Consequently, DTV sampling is pivotal for effective feature selection, directly improving the model’s performance.

\input{table/albation_DTV_Random_sampling}

\subsubsection{Analysis of Offset Concept Transition in MTS}
When designing the JA attention module, we modify the concept of the offset mechanism in deformable attention~\cite{zhu2021deformable} to be more suitable for MTS data.
In other words, we redefine offsets, transforming their role from being sampling points themselves to serving as guidelines for constructing the cross-axis pool.
% In this section, we analyze the justification of this conceptual shift of the offsets through additional experiments.
In this analysis, we justify this conceptual shift of the offsets through additional experiments.

Table~\ref{tbl:ablation_offset} presents the results of the experiment labeled as (a) Point-level, which follows the conventional offset concept used in deformable attention modules.
Row (b) Pattern-level in the table describes the results obtained using the proposed JA attention module.
For a fair comparison,  we ensure that the number of offsets in the point-level experiment is equal to the number of sampling points $K$ in our method.
% we match the number of offsets in the point-level experiment with the number of sampling points $K$ in our method.

The experimental results demonstrate the advantages of the pattern-level approach over the point-level approach in MTS forecasting.
The point-level approach, which focuses on isolated offset points, may exhibit suboptimal performance due to its constrained scope, which limits its ability to capture broader inter-dependencies between variates.
In contrast, the pattern-level approach enables relevant sampling across both the temporal and variate axes.
This method combines precise feature selection within specific temporal and variate regions with the ability to incorporate dynamically significant patterns.
Therefore, the pattern-level approach effectively captures complex inter-variate and temporal dynamics, leading to consistent improvements in forecasting performance.

\input{table/ablation_offset}

\subsubsection{Qualitative Analysis for DTV Sampling} 

\begin{figure}[t]
\begin{subfigure}{0.495\linewidth}
    \centering
    \includegraphics[width=\textwidth]{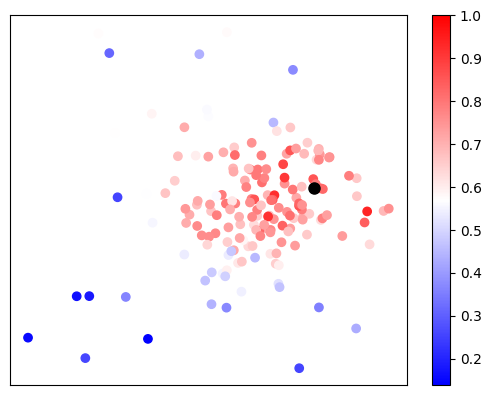}
    \caption{Trend}
    \label{fig5a}
\end{subfigure}
\hfill
\begin{subfigure}{0.495\linewidth}
    \centering
    \includegraphics[width=\textwidth, height=3.3cm]{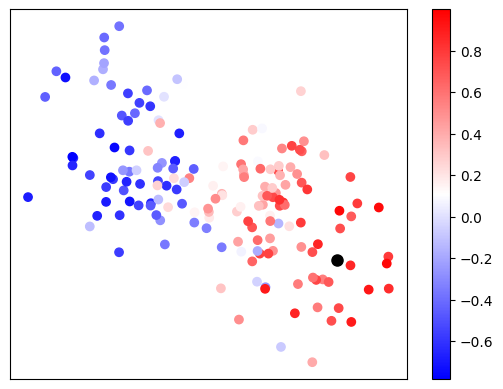}
    \caption{Seasonality}
    \label{fig5b}
\end{subfigure}

\caption{Qualitative analysis for DTV sampling.
The black points represent the query feature, while other features are colored based on their cosine similarity to the query: red for high similarity and blue for low similarity.
(a) and (b) represent 2D embedding spaces for the trend and seasonality components, $X^{Tr}$ and $X^{Se}$, of an input $X$, respectively.}

\label{fig:emb_space}
\end{figure}

We present qualitative results in Fig.~\ref{fig:emb_space} to verify the effect of using the 2D embedding space for DTV sampling.
% These results are obtained for the trend and seasonality components of an input $X$, respectively, using trained TiVaT on the ETTh1 dataset.
These results are obtained for the trend and seasonality components of an input $X$ from the ETTh1 dataset.
% We employ cosine similarity to measure the semantic relevance between the query feature at a reference point and other features in a candidate pool within feature dimension. 
We employ cosine similarity along the dimensional axis to measure the semantic relevance between the query feature at a reference point and other features in the candidate pools.
% In Fig.~\ref{fig:emb_space}, the query feature is displayed in black, and other features are expressed in red as the higher similarity to the query and blue as the lower similarity.
Fig.~\ref{fig:emb_space} provides unified visualizations of the similarity between the query feature and others in its candidate pools, along with their spatial distribution in the 2D embedding space used for DTV sampling.

% As described in Fig.~\ref{fig:emb_space}, the features with higher similarity cluster near the query, while those with lower similarity are positioned farther away. 
As intended, features with higher similarity cluster near the query, while features with lower similarity are placed farther away. 
% First, this result indirectly suggests why sampling should be done, since there are also features with zero or less similarity within the candidate pool, which can be noise information.
This observation supports the DTV sampling strategy, which samples relevant features based on Euclidean distance in the 2D embedding space.
Additionally, supplementary visualizations provided in Supp.~\ref{sec:vis_supp_DTV} showcase examples for randomly selected reference points, ensuring that the analysis is unbiased and not restricted to specific cases.

% \subsubsection{Visualization for Capturing Asynchrony}
%\subsubsection{Analysis for Capturing Asynchronous Interactions}
% \subsubsection{Visualization for Examining Asynchronous Interactions}
\subsubsection{Visualization of Asynchronous Interactions}
\begin{figure}[h]
\begin{subfigure}{0.495\linewidth}
    \centering
    \includegraphics[width=\textwidth]{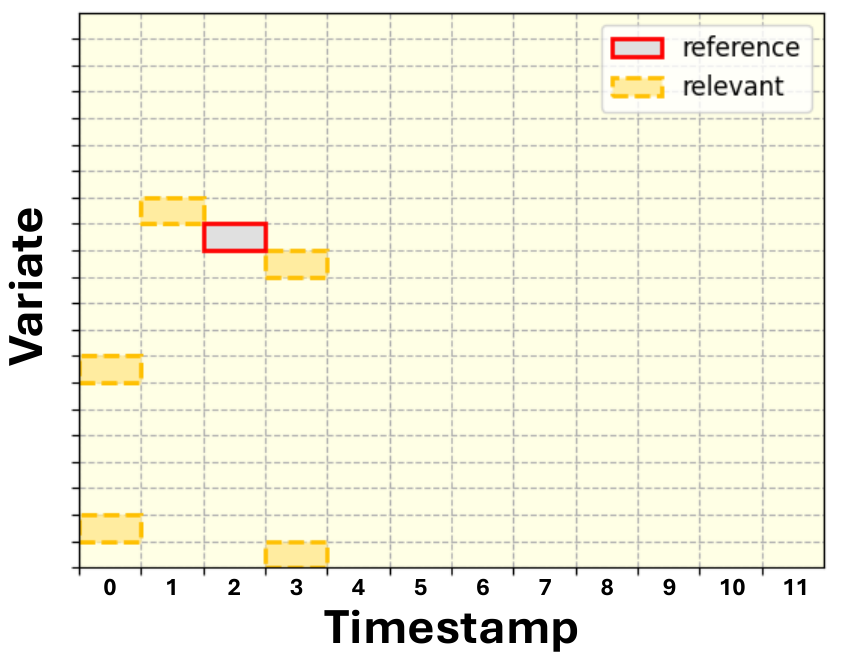}
    % \caption{The 6\textsuperscript{th} timestamp}
    \caption{Timestamp $t_3$}
\end{subfigure}
\hfill
\begin{subfigure}{0.495\linewidth}
    \centering
    \includegraphics[width=\textwidth]{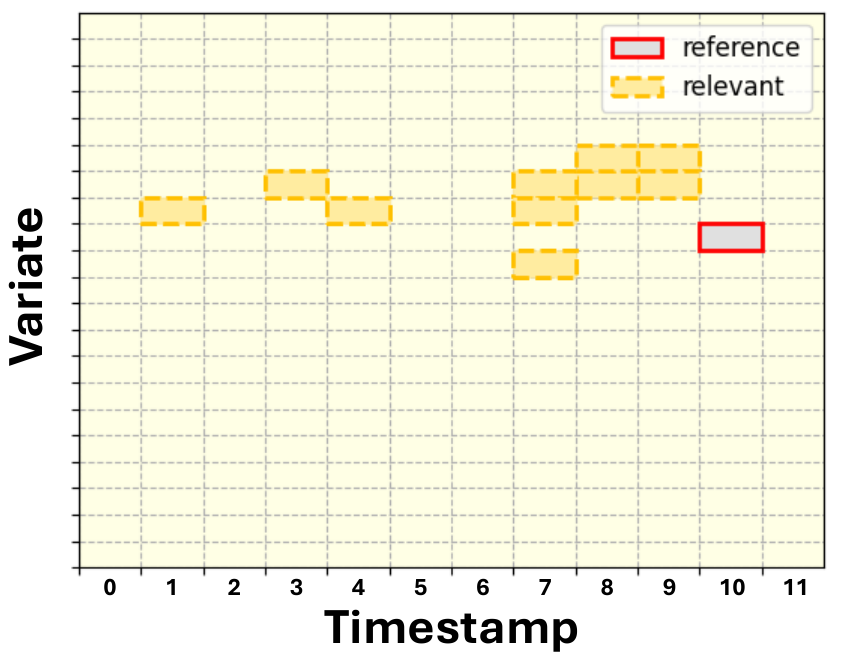}
    % \caption{The 14\textsuperscript{th} timestamp}
    \caption{Timestamp $t_{11}$}
\end{subfigure}
\caption{Visualization of grid maps illustrating a reference point and its relevant points extracted in JA attention module. (a) and (b) describe grid maps for different reference points along timestamps for specific variates in Weather dataset. The red box indicates the reference point and the yellow boxes represent the features strongly related to it across the variate and temporal dimensions.}
\label{fig:vis}
\end{figure}

To examine whether the proposed model, TiVaT, captures asynchronous interactions, we visualized grid maps illustrating the relevant points extracted in the cross-axis pool for a given reference point. 
% To verify whether the proposed model TiVaT actually captures asynchronous interactions, we visualized grid maps illustrating the relevant points extracted in the cross-axis pool for a given reference point. 
% Fig.~\ref{fig:vis} illustrates that for a particular variate in ETTh1, the asynchronous dependencies captured by the model change dynamically over time.
Fig.~\ref{fig:vis} shows how the asynchronous dependencies captured by the model evolve dynamically over time for a particular variate in the Weather dataset.
We provide additional visualizations of variate-specific and time-specific reference points across various datasets in the Supp.~\ref{sec:vis_supp}.
% These visualizations highlight TiVaT's ability to capture diverse asynchronous interactions and its interpretability for a phenomenon such as lead-lag for a variate at a specific timestamp in MTS data.
These visualizations highlight TiVaT's ability to capture diverse asynchronous interactions and demonstrate its interpretability in identifying such interactions for a variate at a specific timestamp in MTS data.

%% file: table/ablation_JA_attention.tex
\begin{table}[t]
\centering

% \scriptsize
\resizebox{\columnwidth}{!}{%
\begin{tabular}{clll|cc|cc|cc}
\hline
\multicolumn{4}{l|}{\multirow{2}{*}{Method}} & \multicolumn{2}{c|}{ETTh1}                   & 
\multicolumn{2}{c|}{Exchange}                & \multicolumn{2}{c}{Weather}                 \\ \cline{5-10} 
\multicolumn{4}{l|}{}                                  & MSE                  & MAE                   & MSE                  & MAE                 & MSE                  & MAE                  \\ \hline
\multicolumn{4}{l|}{(a) Full Attention}                    & \multicolumn{1}{c}{0.456} & \multicolumn{1}{c|}{0.449} &
\multicolumn{1}{c}{0.385} & \multicolumn{1}{c|}{0.413} & \multicolumn{1}{c}{0.270} & \multicolumn{1}{c}{0.289} \\
\multicolumn{4}{l|}{(b) Two-Stage Attention}                 &    0.478                &     0.465            & 0.427                      &    0.438                   &  0.276               &  0.300                   \\
\multicolumn{4}{l|}{(c) JA Attention}                      & \textbf{0.434}            & \textbf{0.435}             & 
\textbf{0.349}            & \textbf{0.398}             & \textbf{0.240}            & \textbf{0.270}            \\
\hline
\end{tabular}%
}
% \caption{Ablation Study on JA Attention. (a) refers to the method that replaces JA Attention blocks with the Transformer encoder, (b) refers to the method that replaces JA Attention blocks with the Crossforemr encoder, and (c) refers to our proposed method.}
\caption{Ablation on JA attention. (a) replaces JA attention blocks with the vanilla Transformer encoder, (b) replaces them with the Crossformer encoder, and (c) utilizes our JA attention module.}
\label{tbl:ablation_JA_attn}
\end{table}

%% file: table/albation_DTV_Random_sampling.tex
% \begin{table}[h]
% \centering

% % \scriptsize
% \resizebox{\columnwidth}{!}{%
% \begin{tabular}{cc|cc|cc|cc}
% \hline
% \multicolumn{2}{c|}{Sampling Method} & \multicolumn{2}{c|}{ETTh1} & \multicolumn{2}{c|}{Exchange} & \multicolumn{2}{c}{Weather} \\
% \hline
% Random & DTV & MSE & MAE & MSE & MAE & MSE & MAE \\
% \hline
% \checkmark & & 0.455 & 0.442 & 0.361 & 0.407 & 0.243 & 0.273 \\
% & \checkmark & \textbf{0.434} & \textbf{0.435} & \textbf{0.349} & \textbf{0.398} & \textbf{0.240} & \textbf{0.270} \\
% \hline
% \end{tabular}%
% }
% \caption{Ablation Study on DTV Sampling. A check mark (\checkmark) indicates that a particular sampling method was used. Random sampling refers to the process of selecting 
% $K$ features randomly from the candidate pools.}
% \label{tbl:ablation_DTV}
% \end{table}

% \begin{table}[h]
% \centering

% % \scriptsize
% \resizebox{\columnwidth}{!}{%
% \begin{tabular}{cc|cc|cc|cc}
% \hline
% \multicolumn{2}{l|}{\multirow{2}{*}{Sampling method}} & \multicolumn{2}{c|}{ETTh1} & \multicolumn{2}{c|}{Exchange} & \multicolumn{2}{c}{Weather} \\
% \hline
%  & &  MSE & MAE & MSE & MAE & MSE & MAE \\
% \hline

% \checkmark & & 0.455 & 0.442 & 0.361 & 0.407 & 0.243 & 0.273 \\
% & \checkmark & \textbf{0.434} & \textbf{0.435} & \textbf{0.349} & \textbf{0.398} & \textbf{0.240} & \textbf{0.270} \\
% \hline
% \end{tabular}%
% }

\begin{table}[t]
\resizebox{\columnwidth}{!}{%
\begin{tabular}{cll|cc|cc|cc}
\hline
\multicolumn{3}{l|}{\multirow{2}{*}{Sampling method}} & \multicolumn{2}{c|}{ETTh1}      & \multicolumn{2}{c|}{Exchange}      & \multicolumn{2}{c}{Weather}     \\ \cline{4-9} 
\multicolumn{3}{l|}{}                                 & MSE            & MAE            & MSE            & MAE            & MSE            & MAE            \\ \hline

\multicolumn{3}{l|}{(a) w/o Sampling}    & {0.454} & {0.443}  & {0.380} & {0.410}&{0.246} & {0.275}\\ % \hline
\multicolumn{3}{l|}{(b) Random Sampling}    & 0.455          & 0.442          &    0.361       & 0.407  & 0.243          & 0.273          \\ 
\multicolumn{3}{l|}{(c) DTV Sampling}  & \textbf{0.434} & \textbf{0.435} & \textbf{0.349}  & \textbf{0.398} & \textbf{0.240} & \textbf{0.270} \\ \hline
\end{tabular}
}
\caption{Ablation on DTV sampling. (a) utilizes all points from the candidate pools, (b) randomly selects $K$ features from the candidate pools, and (c) employs the proposed DTV sampling.}
\label{tbl:ablation_DTV}
\end{table}

%% file: table/ablation_offset.tex
\begin{table}[t]
\centering
\resizebox{\columnwidth}{!}{%
\begin{tabular}{clll|cc|cc|cc}
\hline
\multicolumn{4}{c|}{\multirow{2}{*}{Offset Concept}} & \multicolumn{2}{c|}{ETTh1}                   &
\multicolumn{2}{c|}{Exchange}                & \multicolumn{2}{c}{Weather}                 \\ \cline{5-10} 
\multicolumn{4}{c|}{}                           & MSE                  & MAE   & MSE                  & MAE     & MSE                  & MAE                \\ \hline
\multicolumn{4}{l|}{(a) Point-level}                     &   0.451                   & 0.439   &      0.402              &        0.424               &           0.246           &   0.274                   \\

\multicolumn{4}{l|}{(b) Pattern-level}     & \textbf{0.434}            & \textbf{0.435}             & \textbf{0.349}            & \textbf{0.398}             & \textbf{0.240}            & \textbf{0.270}            \\ \hline
\end{tabular}
}
\caption{Analysis of offset concept transition in MTS. (a) refers to using the offsets only the relevant points, while (b) uses them as the guidelines for DTV sampling.}
\label{tbl:ablation_offset}
\end{table}

%% file: 5_Conclusion.tex
\section{Conclusion}
\label{con}
In this work, we present TiVaT, the first Transformer-based CD model for MTS forecasting that employs a single unified module to capture asynchronous dependencies.
Unlike existing Transformer-based CD models, which often overlook lead-lag dynamics, TiVaT leverages its JA attention mechanism to jointly model temporal and variate dependencies, addressing the complex relationships inherent in MTS.
Furthermore, DTV sampling enhances TiVaT’s capability by extracting key patterns through a learned 2D embedding space, effectively reducing noise and improving forecasting accuracy.
Extensive experiments on diverse benchmark datasets demonstrate that TiVaT achieves overall performance compared to SOTA models.
By addressing critical challenges in MTS forecasting, particularly modeling of asynchronous dependencies, TiVaT establishes itself as a robust framework for managing complex relationships and interactions inherent in real-world datasets. 
We believe this study is pioneering in its approach to simultaneously modeling temporal and inter-variate dependencies, serving as a catalyst for shaping new directions in MTS forecasting.

%% file: 6_Appendix.tex
\appendix

\section{Experiment Details}

\subsection{Datasets}
\label{sec:supp_data}

We conduct experiments on eight real-world datasets to evaluate the performance of the proposed TiVaT, which include the following:

\begin{itemize}
    \item ETT (ETTh1, ETTh2, ETTm1, ETTm2)~\cite{zhou2021informer}: Electricity transformer data with 7 factors, including hourly (ETTh1/ETTh2) and 15-minute (ETTm1/ETTm2) records, from July 2016 to July 2018.
    \item Exchange~\cite{wu2021autoformer}: Daily exchange rate data from eight countries, spanning 1990 to 2016.
    \item Weather~\cite{wu2021autoformer}: Meteorological data with 21 factors recorded every 10 minutes in 2020 by the Max Planck Biogeochemistry Institute.
    \item ECL~\cite{wu2021autoformer}: Hourly electricity consumption data for 321 clients, covering 2012 to 2014.
    \item Traffic~\cite{wu2021autoformer}: Hourly road occupancy rates measured by 862 sensors in California from January 2015 to December 2016.
\end{itemize}

We adopt the data processing steps and the train-validation-test splitting method described in iTransformer~\cite{liu2024itransformer}.
The datasets for training, validation, and testing are strictly separated in chronological order to prevent any potential data leakage.
The datasets are normalized to a standard normal distribution using the mean and standard deviation from the training set.
The details of datasets are provided in Table~\ref{tbl:dataset_info}.

\input{table/appendix_dataset}
 
\subsection{Training Configuration}
\label{sec:supp_training}

% Our experiments are conducted using PyTorch~\cite{paszke2017automatic} on multiple NVIDIA A100 GPUs, each with 80GB of memory.
The ADAM optimizer~\cite{kingma2015adam} is employed to optimize the $L_2$ loss, with the initial learning rate selected from \{$10^{-3}$, $5 \times 10^{-4}$, $10^{-4}$, $5 \times 10^{-5}$, $10^{-5}$\}.
% We also apply a learning rate scheduler, either step-based or cosine annealing~\cite{loshchilov2017sgdr}, for optimization.
A learning rate scheduler, either step-based or cosine annealing~\cite{loshchilov2017sgdr}, is applied for optimization.

For the forecasting setup, we use a fixed lookback window $L_H$ of 96 time steps for the ETT, Exchange, Weather, ECL, and Traffic datasets, with prediction horizons $L_F \in \{96, 192, 336, 720\}$.

\paragraph{Hyperparameter optimization.}

The batch size was selected from \{4, 8, 16, 32, 64\} across all experimental configurations.
The number of JA attention blocks in the model was varied, with values chosen from the \{2, 3, 4\}.
The dimension of the series representation was selected from \{128, 256, 512, 1024\}.
For each JA attention Block, the percentage parameters for the temporal axis ($p_t$) and variate axis ($p_v$) were set within the range of 0.1 to 0.8.
Additionally, the number of samples for both the self-axis pool ($K^{self}$) and the cross-axis pool ($K^{cross}$) was chosen from \{10, 20, 30, 40, 60, 80\}.

\section{Further Analysis}

% \subsection{Qualitative Results of Guidelines from JA Attention Module}

% When performing an attention operation with a specific time and variate, irrelevant information from other times and variates can act as noise.
% Since no prior study used vanilla cross-attention in this context, we visualized the correlation using a heatmap, as shown in Fig.~\ref{fig:heatmap}.
% The results reveal that certain variates at different times are irrelevant and can act as noise. In contrast, the guidelines extracted using TiVaT's JA attention effectively capture the most important information.

% \begin{figure}[!h]
% \begin{subfigure}{0.49\linewidth}
%     \centering
%     \includegraphics[width=\textwidth]{figs/trend_heatmap.png}
%     \caption{}
%     \label{fig4a}
% \end{subfigure}
% \hfill
% \begin{subfigure}{0.49\linewidth}
%     \centering
%     \includegraphics[width=\textwidth]{figs/season_heatmap.png}
%     \caption{}
%     \label{fig4b}
% \end{subfigure}
% \caption{The heatmap uses a white-to-red gradient to indicate the correlation with the reference point, where white represents low correlation and red represents high correlation.
% The dark gray box border highlights the corresponding query, while the light gray box border marks the guideline extracted through TiVaT's JA attention.
% (a) Heatmap and guideline in Trend. (b) Heatmap and guideline in Seasonality.}
% \label{fig:heatmap}
% \end{figure}
% \vspace{10mm}
\subsection{Ablation on Separate Sampling}
\label{sec:supp_sep}
We explore the impact of sampling strategies on capturing complex dependencies in MTS data by comparing the proposed separate sampling (b) with common sampling (a).
The common sampling method (a) conducts DTV sampling only once for a single pool, which is a combination of the self- and cross-axis pools.
The separate sampling performs DTV sampling independently for the self- and cross-axis pools to fully reflect their unique contributions.

The results in Table~\ref{tbl:ablation_separate_sampling} highlight the effectiveness of separate sampling and the importance of self- and cross-axis features.
Self-axis features capture relationships between variates at the same timestep and temporal dependencies within the same variates, while cross-axis features reflect asynchronous interactions between variates across timesteps.
By considering these pools separately, the model can more effectively identify and utilize the diverse relationships inherent in MTS data.

\input{table/ablation_separate_sampling}

\input{table/appendix_ablation_residual_conn}

\subsection{Ablation on Residual Connections in Time Series Decomposition}

% This subsection explores the effect of residual connections on preserving and enhancing temporal patterns in time series decomposition.
% % We compare four configurations: (a) no residual connections, (b) residual connections applied only to the trend component, (c) residual connections applied only to the seasonal component, and (d) residual connections applied to both components. (caption과 중복)
% As shown in Table~\ref{tbl:residual_conn}, incorporating residual connections consistently improves performance across all datasets.
% % The best results are achieved when residual connections are applied to both the trend and seasonal components, highlighting their ability to enhance the model’s capacity to capture complex temporal patterns, leading to more accurate forecasting.
% % The best results are achieved when residual connections are applied to both the trend and seasonal components, as they enhance the model's capacity and improve forecasting performance.

We investigate the impact of residual connections on preserving and enhancing temporal patterns in time series decomposition.
As shown in Table~\ref{tbl:residual_conn}, incorporating residual connections consistently improves performance across all datasets.
This improvement can be attributed to the stabilizing effect of residual connections, which mitigate vanishing gradients and ensure the preservation of essential temporal features.
These findings underscore the critical role of residual connections in maintaining both stability and efficiency during the learning process in MTS forecasting.

\section{Qualitative Analysis for DTV Sampling}
\label{sec:vis_supp_DTV}
We provide additional visualizations from the ETTh1 dataset, showcasing the similarity between the query feature and others in its candidate pools, along with their spatial distribution in the 2D embedding space used for DTV sampling. To ensure the analysis remains objective and avoids potential biases such as cherry-picking, the reference points for these examples were selected randomly. 
Through visualizations, we confirmed that DTV sampling reliably identifies semantically relevant features across a range of scenarios.

%\section{Visualization of Examining Asynchronous Interactions}
\section{Visualization of Asynchronous Interactions}
\label{sec:vis_supp}

\begin{figure*}[!p]
\begin{subfigure}{\linewidth}
    \centering
    \includegraphics[width=0.33\textwidth]{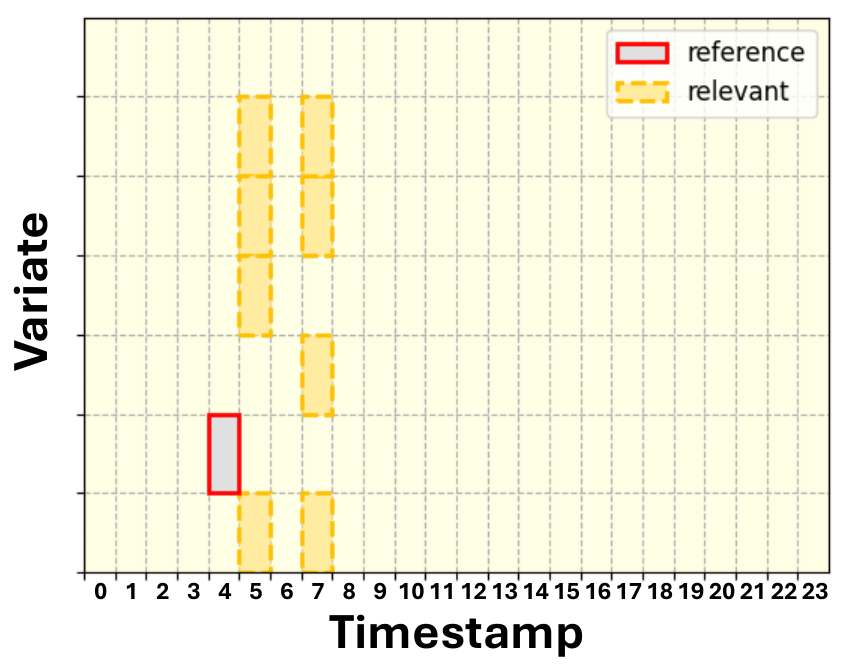}
    \includegraphics[width=0.33\textwidth]{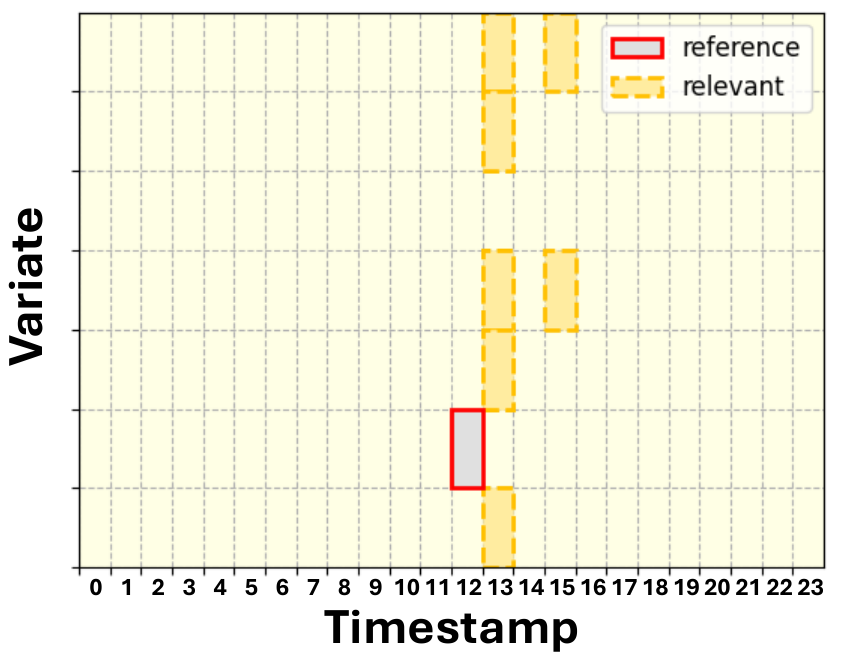}
    \includegraphics[width=0.33\textwidth]{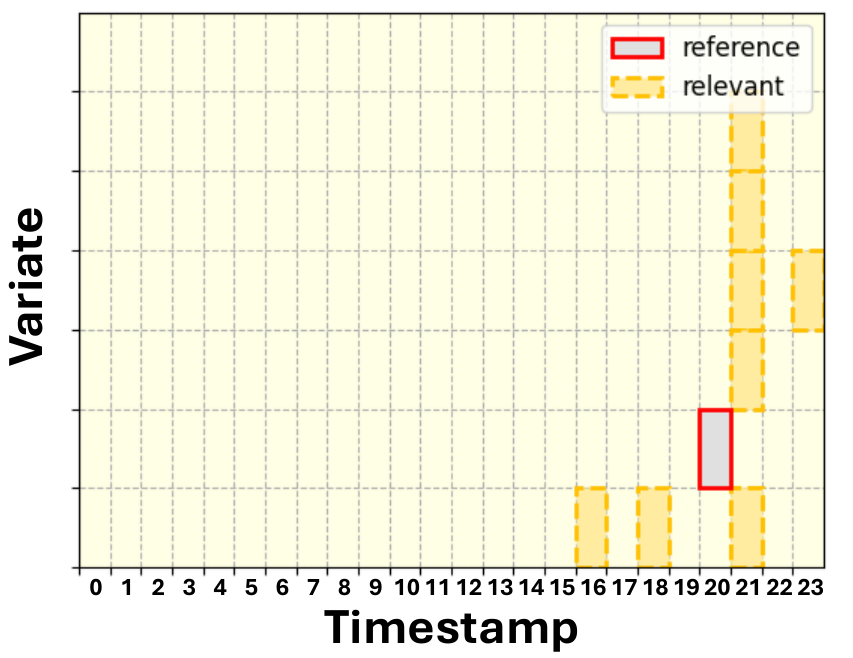}
    \caption{}
\end{subfigure}
\begin{subfigure}{\linewidth}
    \centering
    \includegraphics[width=0.33\textwidth]{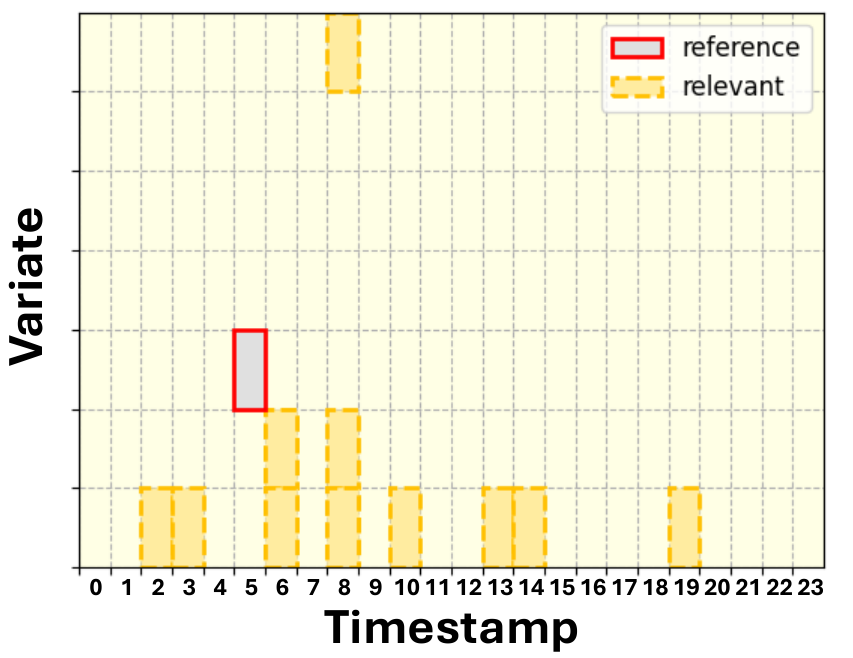}
    \includegraphics[width=0.33\textwidth]{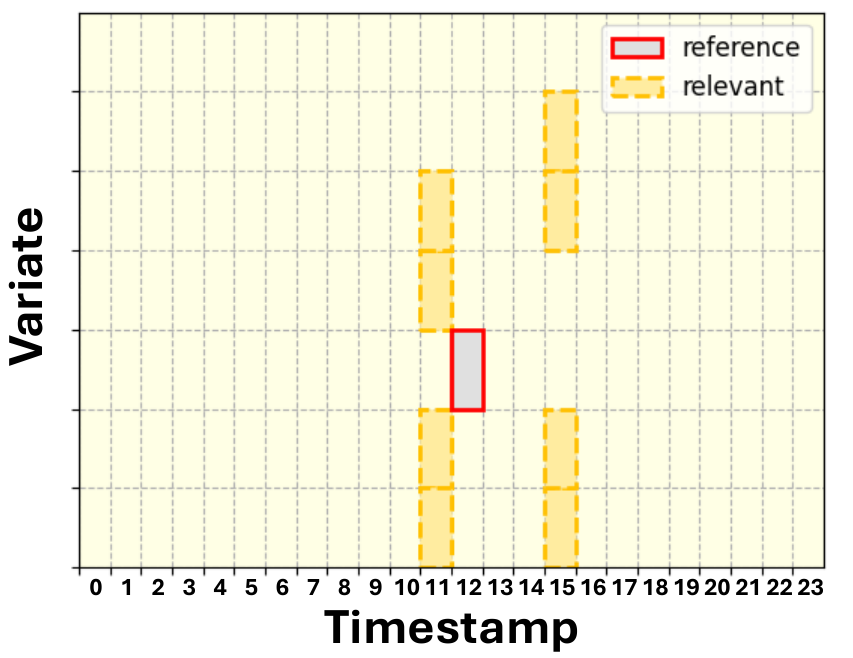}
    \includegraphics[width=0.33\textwidth]{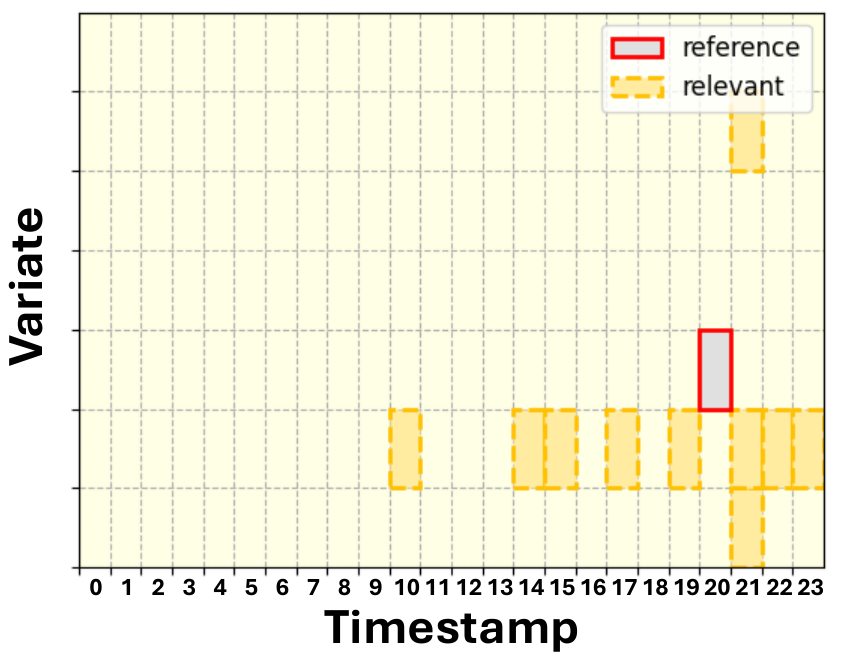}
    \caption{}
\end{subfigure}
\begin{subfigure}{\linewidth}
    \centering
    \includegraphics[width=0.33\textwidth]{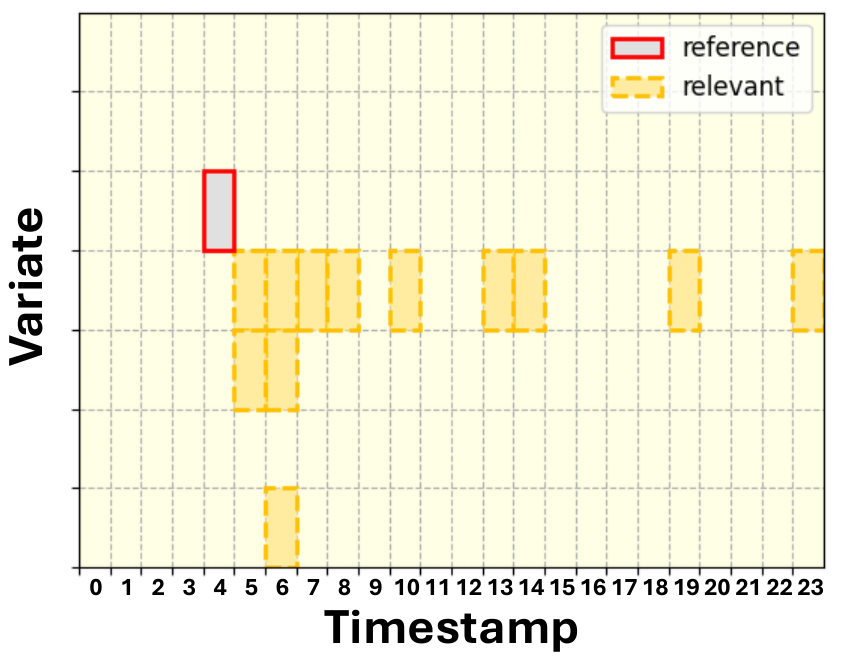}
    \includegraphics[width=0.33\textwidth]{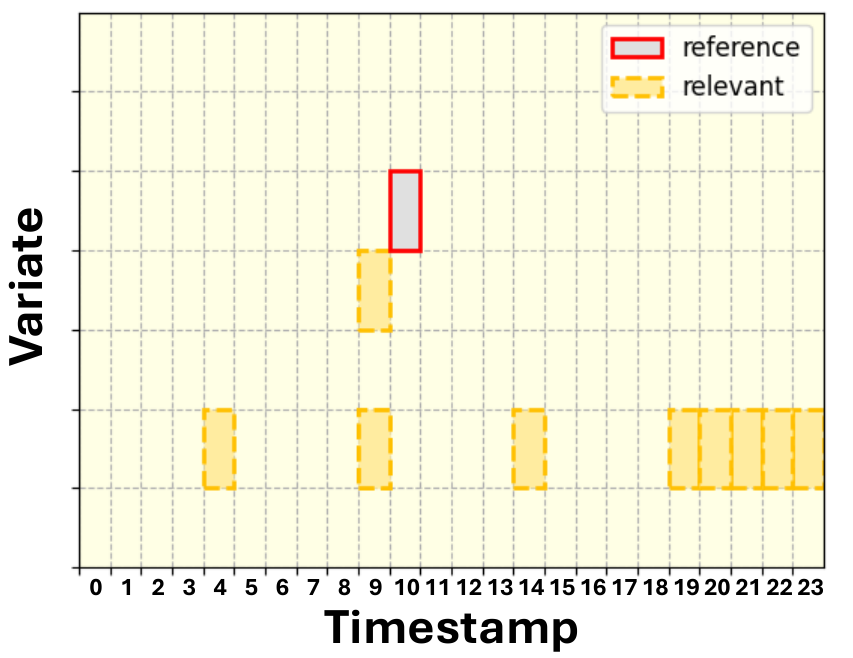}
    \includegraphics[width=0.33\textwidth]{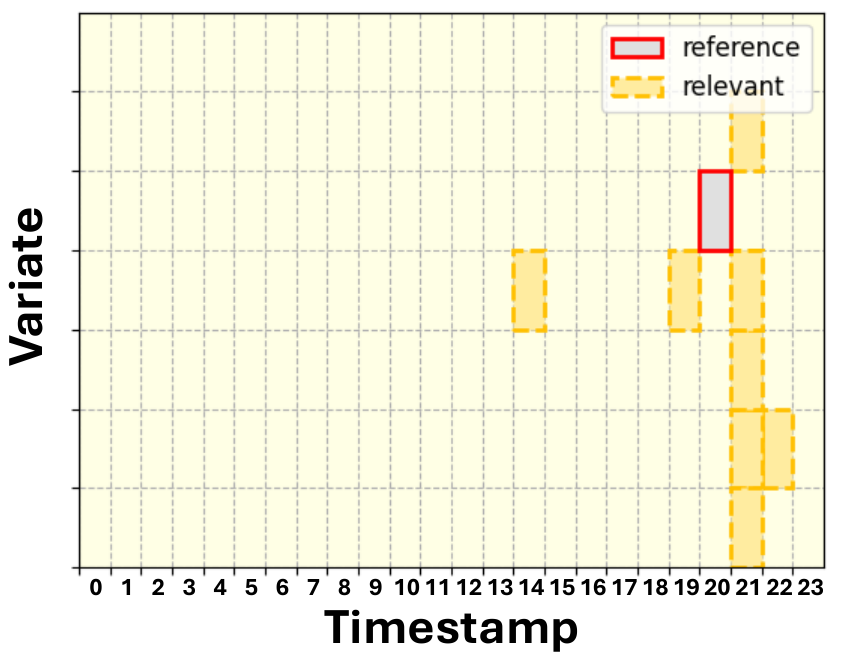}
    \caption{}
\end{subfigure}
\begin{subfigure}{\linewidth}
    \centering
    \includegraphics[width=0.33\textwidth]{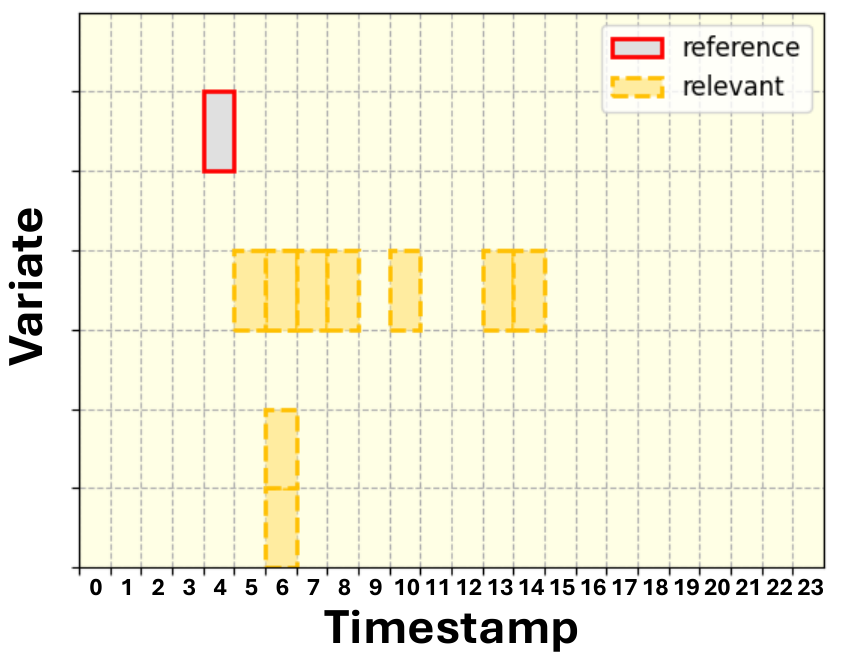}
    \includegraphics[width=0.33\textwidth]{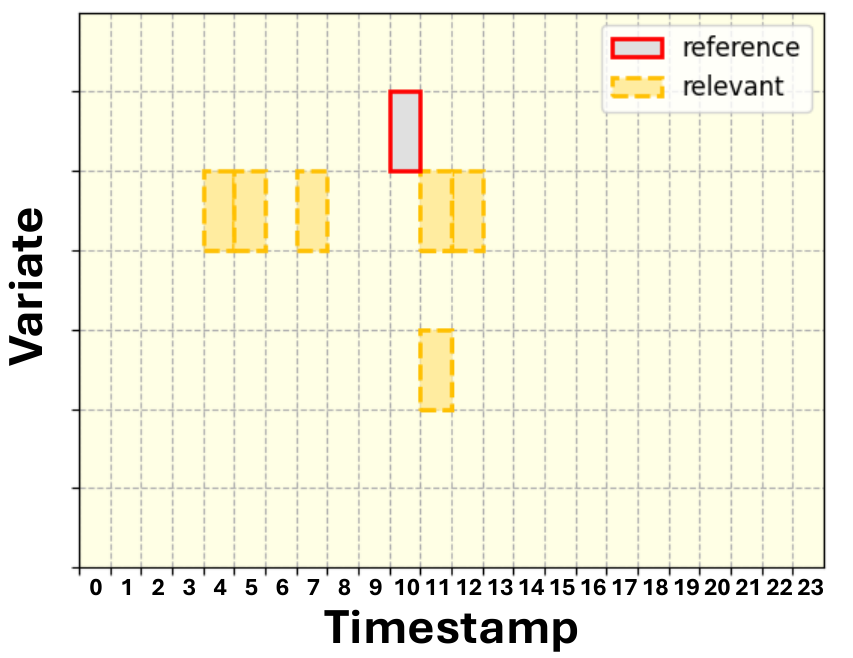}
    \includegraphics[width=0.33\textwidth]{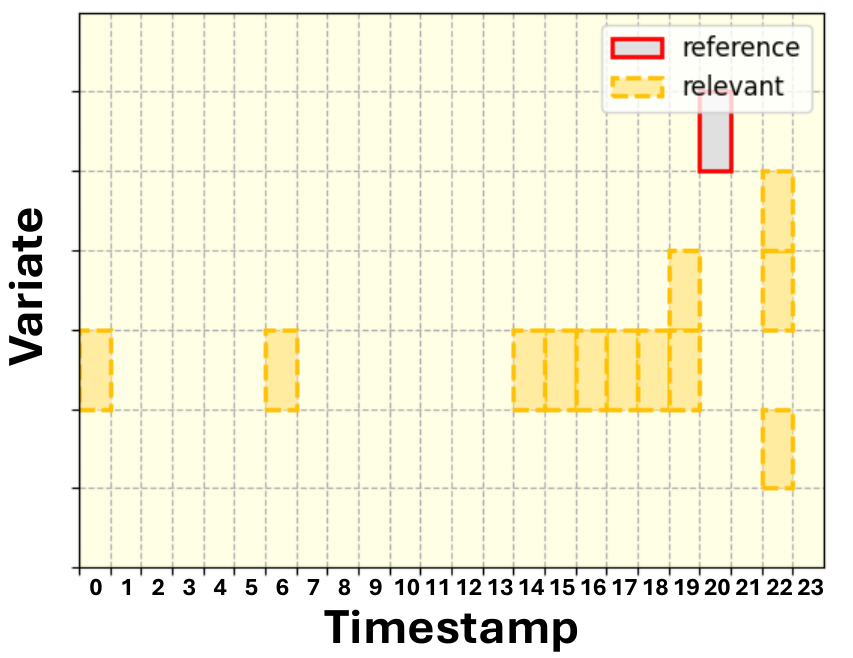}
    \caption{}
\end{subfigure}
\caption{Visualization of asynchronous interactions between the reference point and the relevant points from the ETTh1 dataset.}
\label{fig:heatmap_etth1}
\end{figure*}

\begin{figure*}[!h]
\begin{subfigure}{\linewidth}
    \centering
    \includegraphics[width=0.33\textwidth]{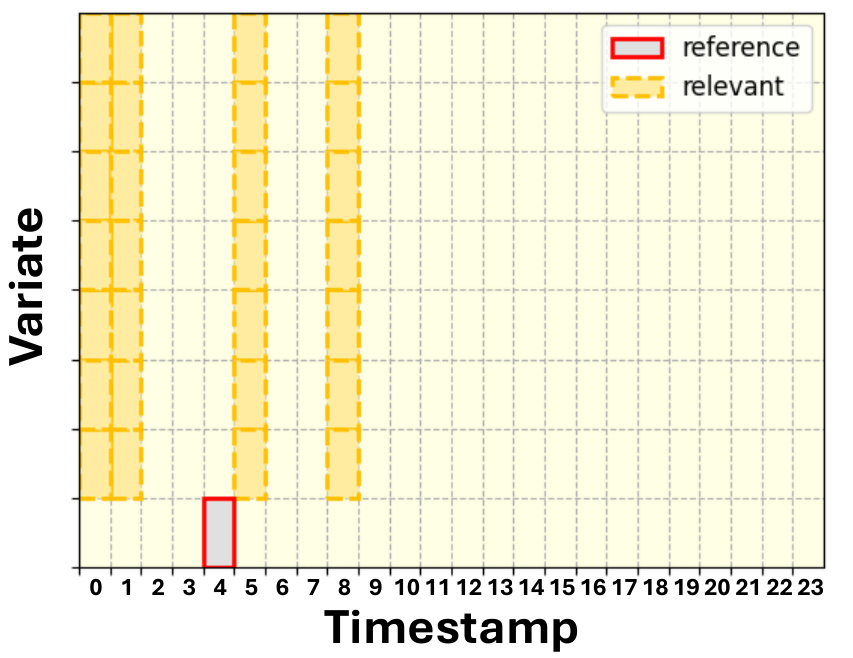}
    \includegraphics[width=0.33\textwidth]{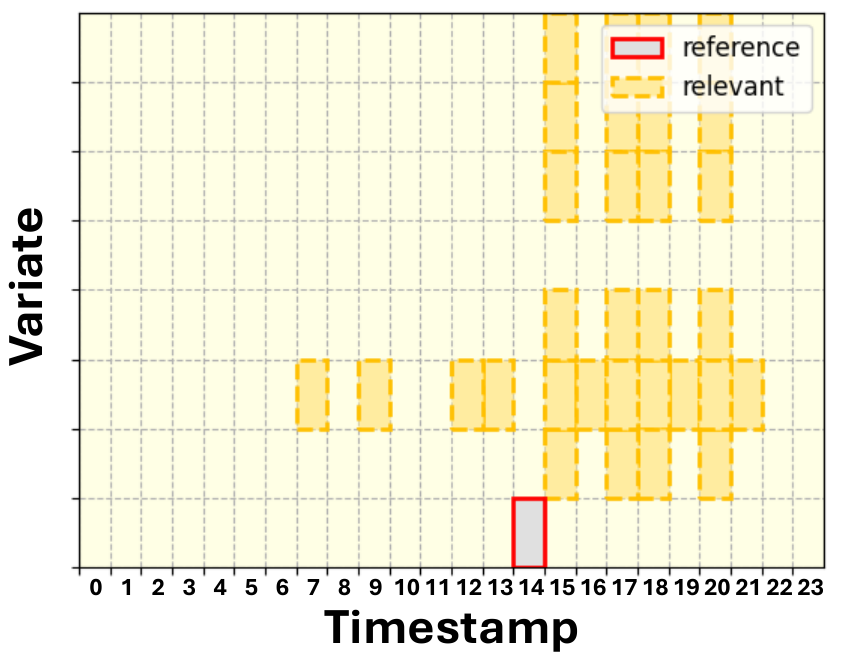}
    \includegraphics[width=0.33\textwidth]{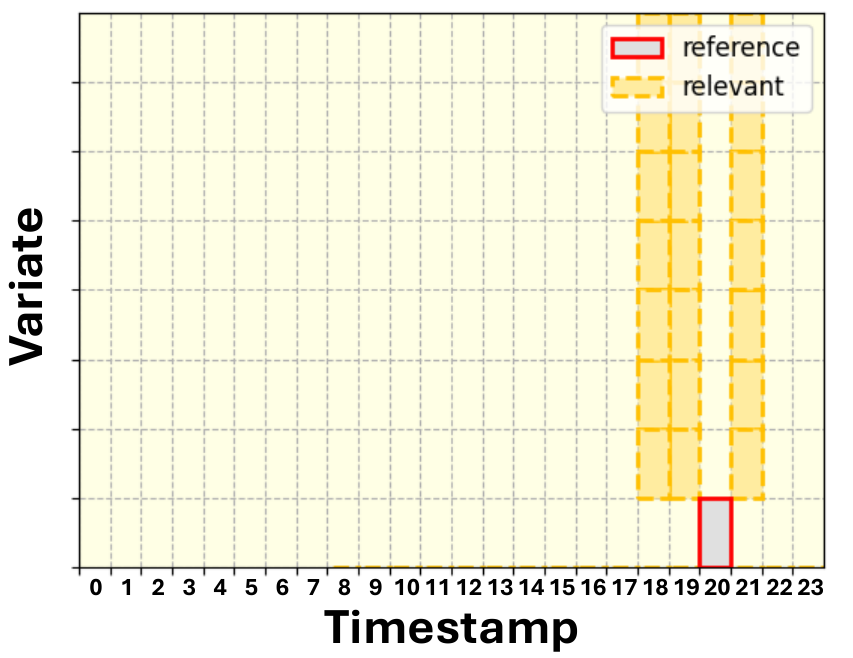}
    \caption{}
\end{subfigure}
\begin{subfigure}{\linewidth}
    \centering
    \includegraphics[width=0.33\textwidth]{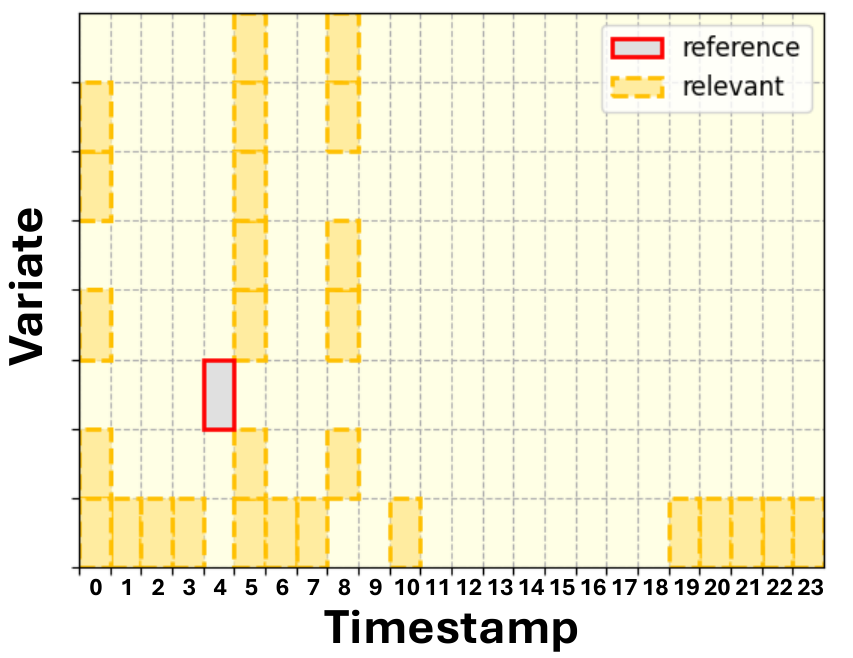}
    \includegraphics[width=0.33\textwidth]{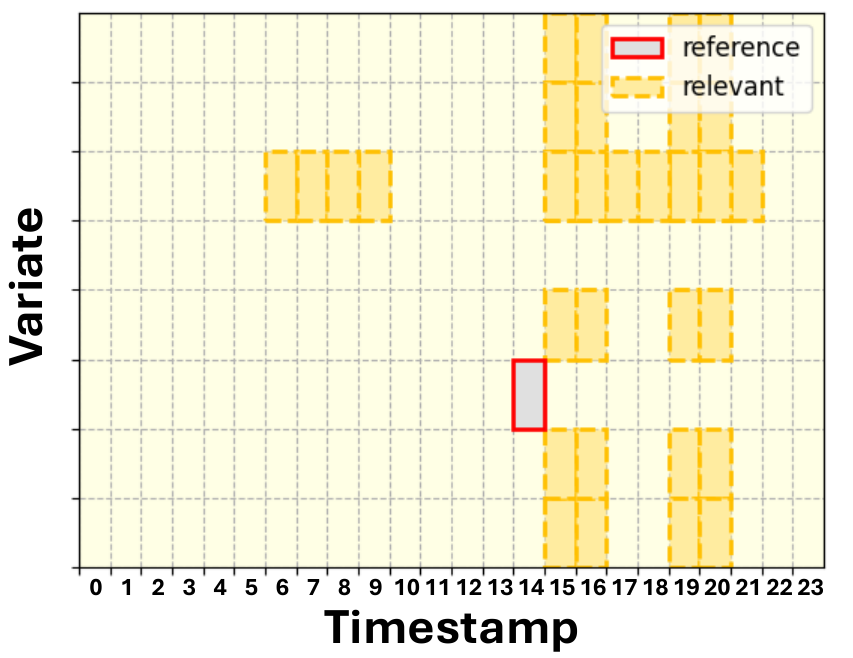}
    \includegraphics[width=0.33\textwidth]{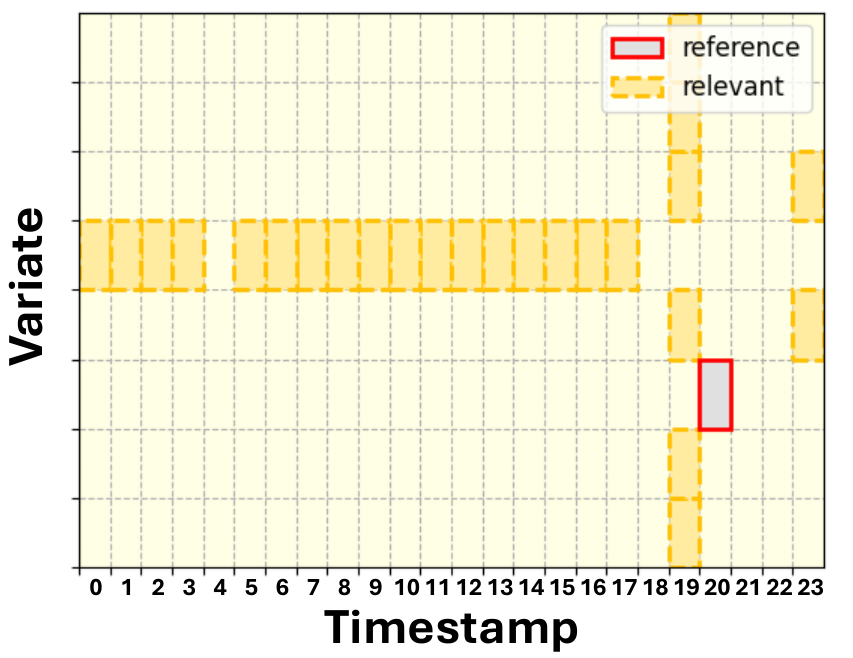}
    \caption{}
\end{subfigure}
\begin{subfigure}{\linewidth}
    \centering
    \includegraphics[width=0.33\textwidth]{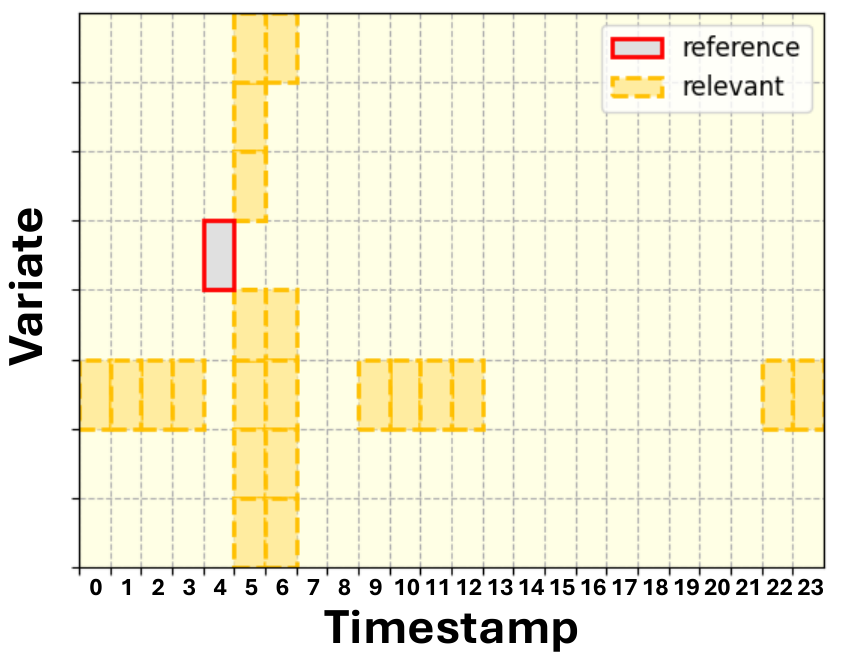}
    \includegraphics[width=0.33\textwidth]{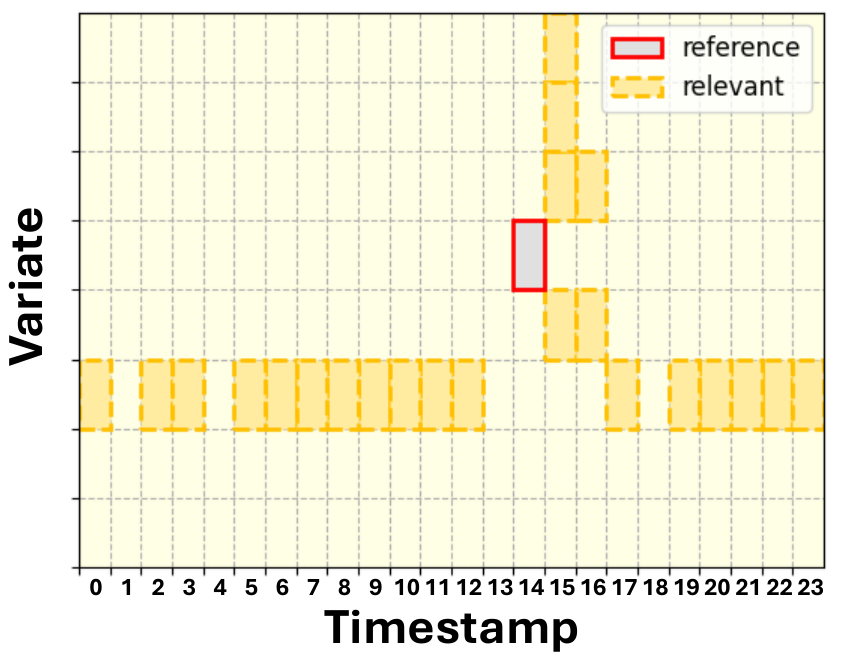}
    \includegraphics[width=0.33\textwidth]{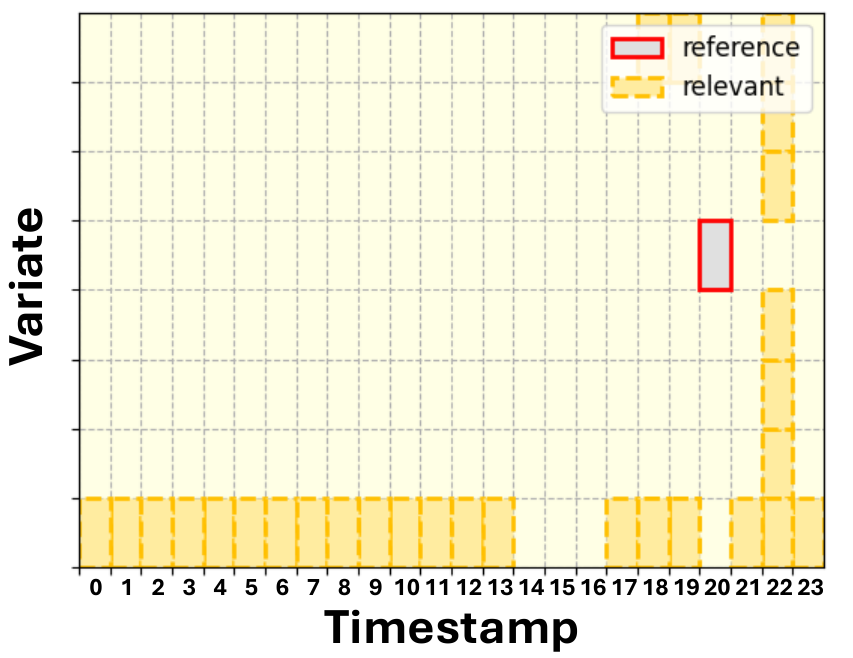}
    \caption{}
\end{subfigure}
\begin{subfigure}{\linewidth}
    \centering
    \includegraphics[width=0.33\textwidth]{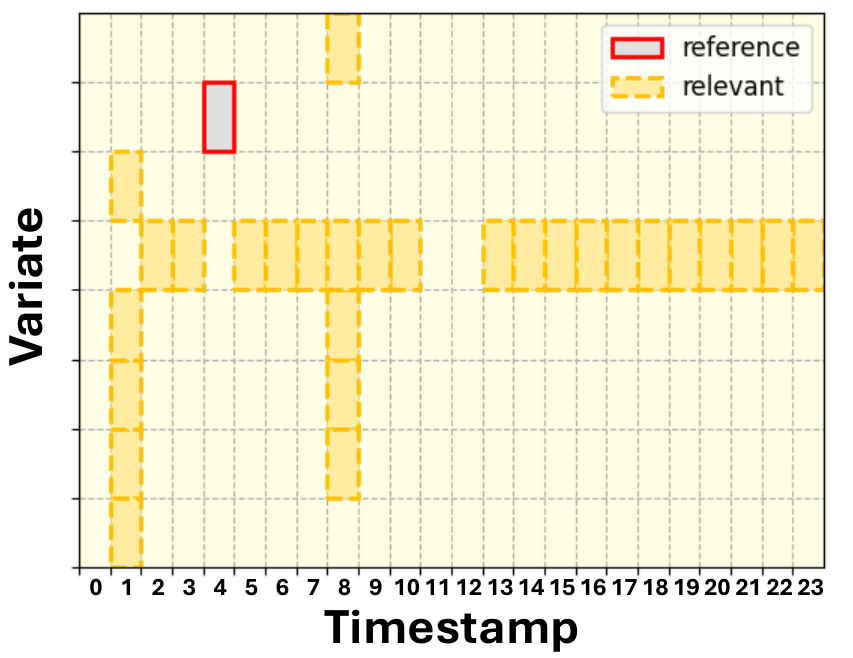}
    \includegraphics[width=0.33\textwidth]{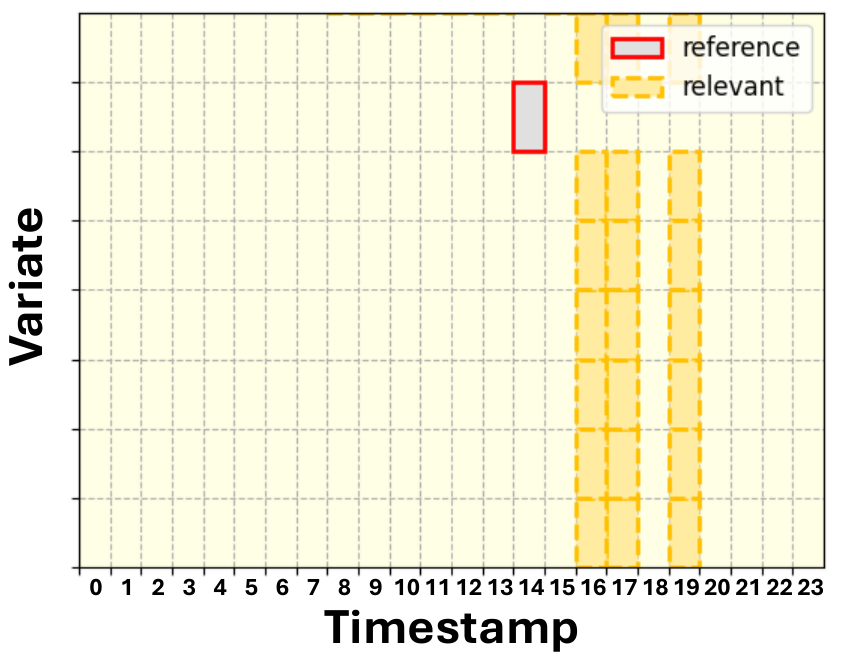}
    \includegraphics[width=0.33\textwidth]{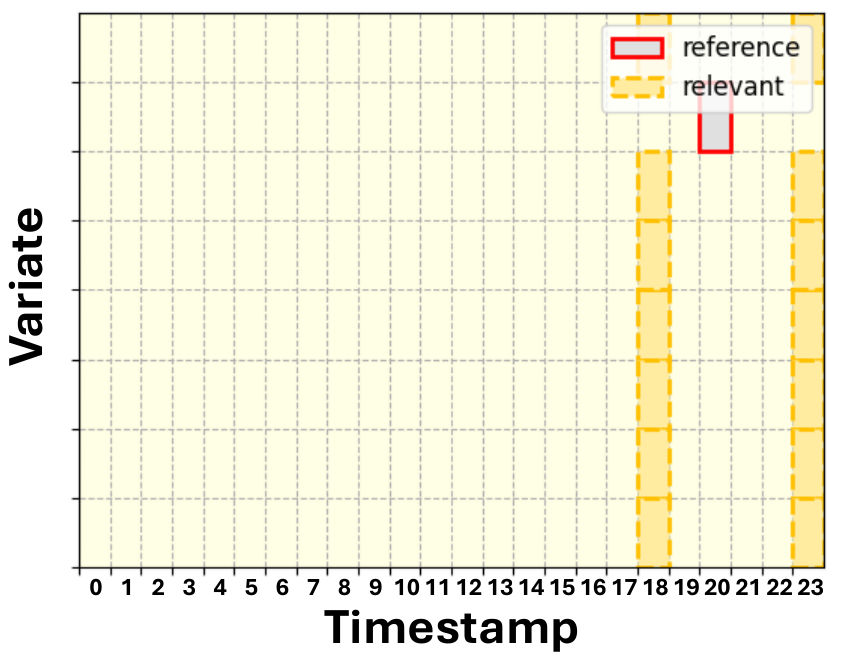}
    \caption{}
\end{subfigure}
\caption{Visualization of asynchronous interactions between the reference point and the relevant points from the Exchange dataset.}
\label{fig:heatmap_exchange}
\end{figure*}

\begin{figure*}[!p]
\begin{subfigure}{\linewidth}
    \centering
    \includegraphics[width=0.33\textwidth]{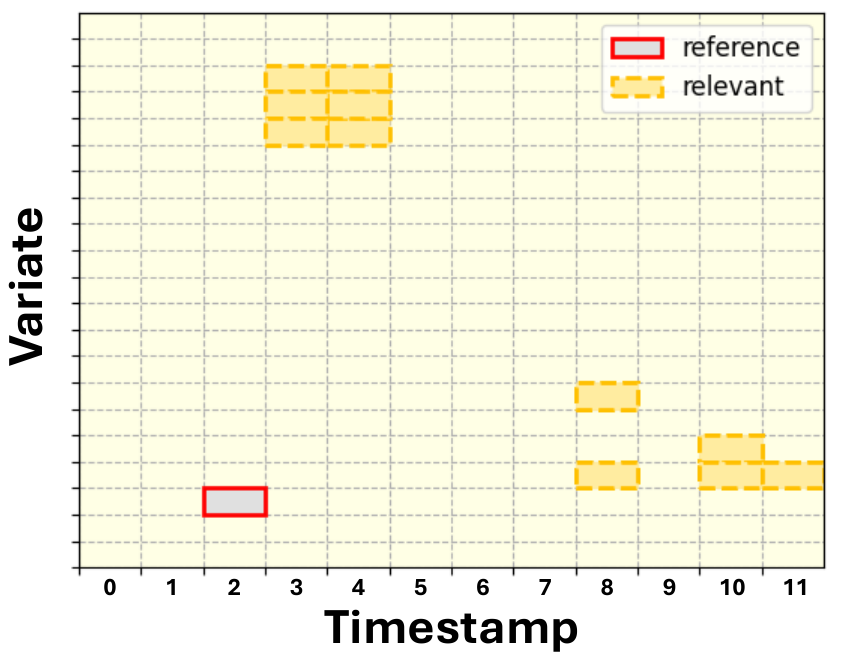}
    \includegraphics[width=0.33\textwidth]{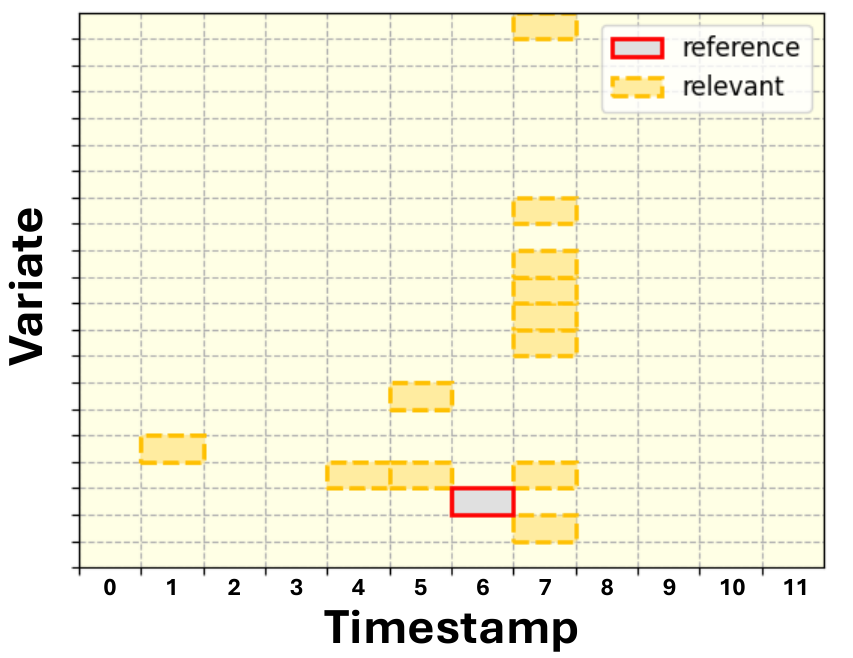}
    \includegraphics[width=0.33\textwidth]{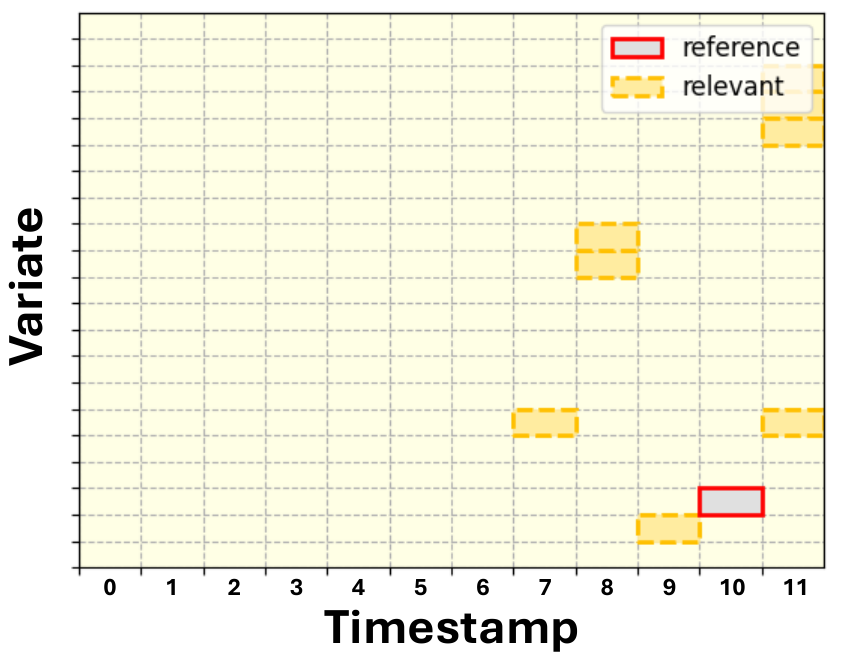}
    \caption{}
\end{subfigure}
\begin{subfigure}{\linewidth}
    \centering
    \includegraphics[width=0.33\textwidth]{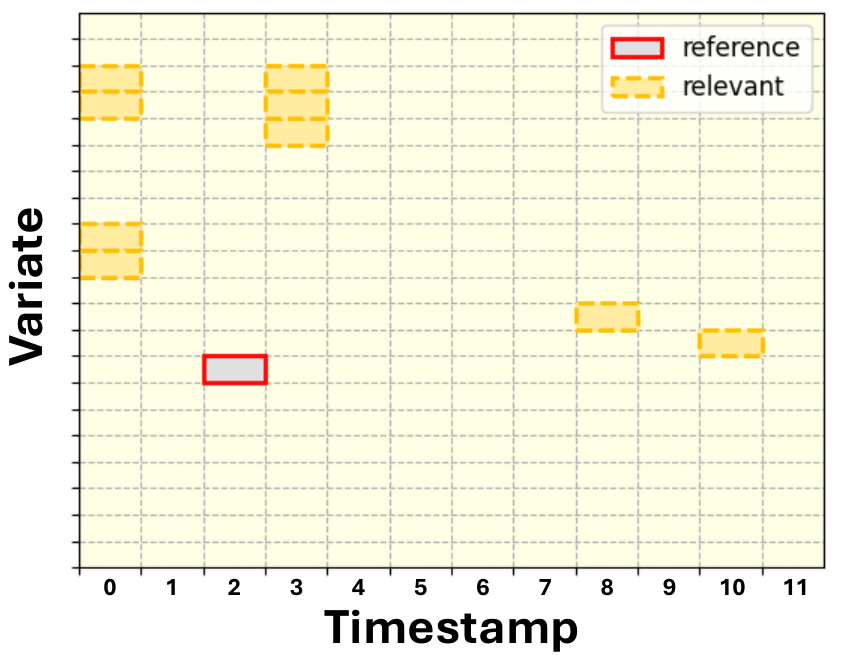}
    \includegraphics[width=0.33\textwidth]{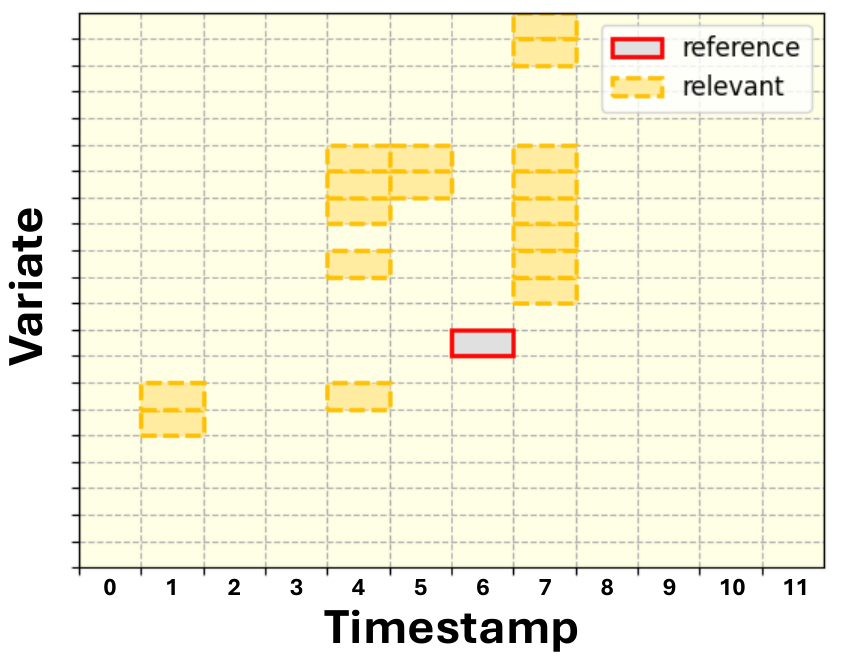}
    \includegraphics[width=0.33\textwidth]{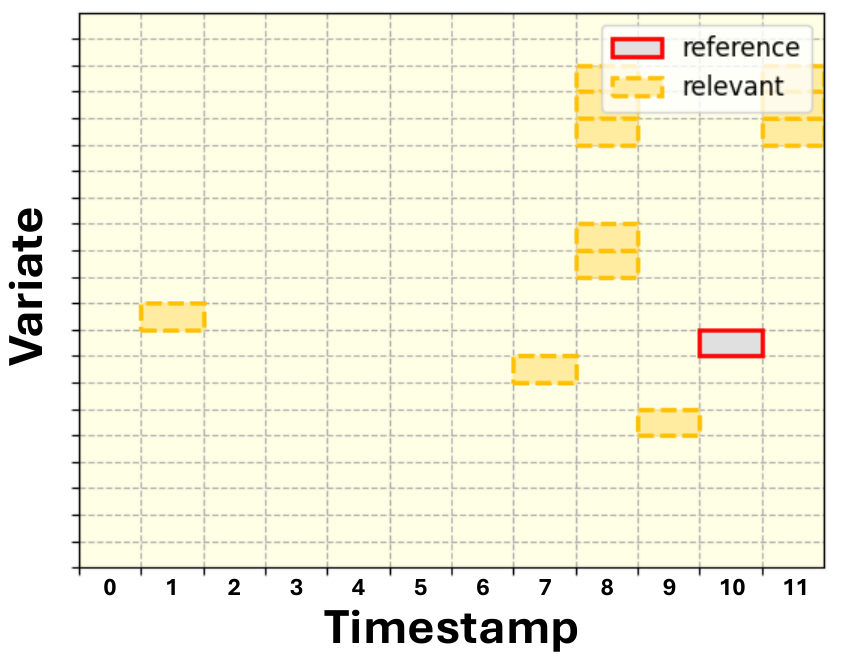}
    \caption{}
\end{subfigure}
\begin{subfigure}{\linewidth}
    \centering
    \includegraphics[width=0.33\textwidth]{figs/Weather_12,2.pdf}
    \includegraphics[width=0.33\textwidth]{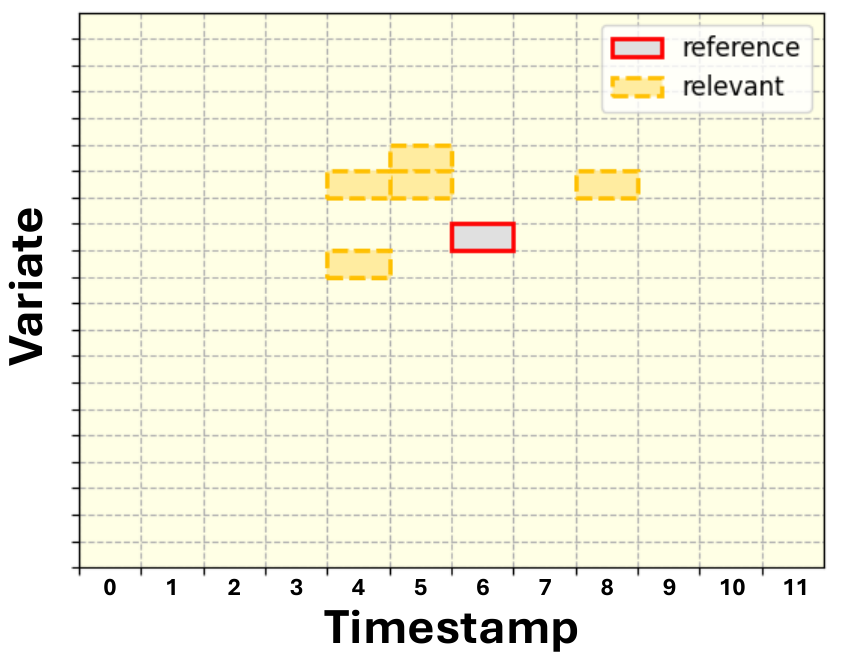}
    \includegraphics[width=0.33\textwidth]{figs/Weather_12,10.pdf}
    \caption{}
\end{subfigure}
\begin{subfigure}{\linewidth}
    \centering
    \includegraphics[width=0.33\textwidth]{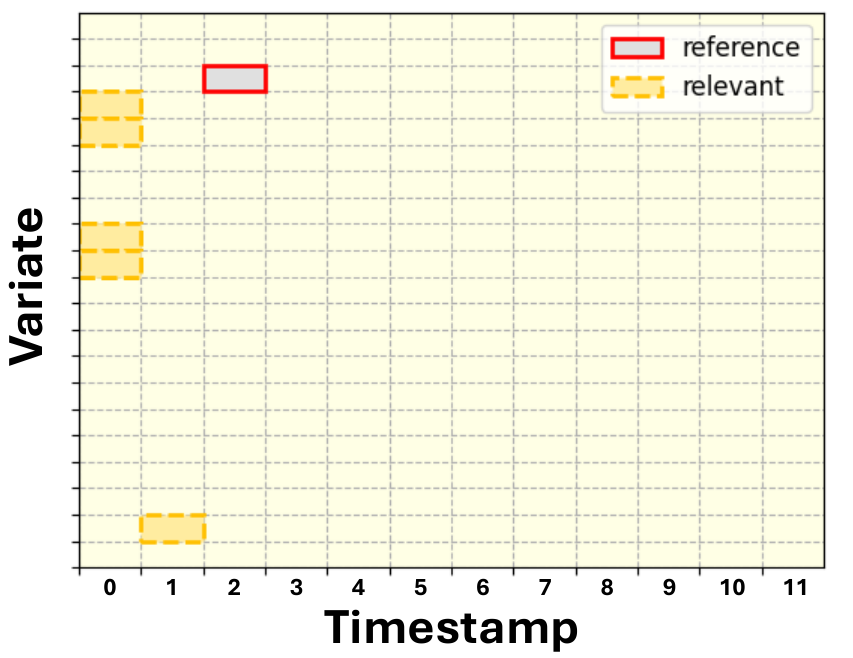}
    \includegraphics[width=0.33\textwidth]{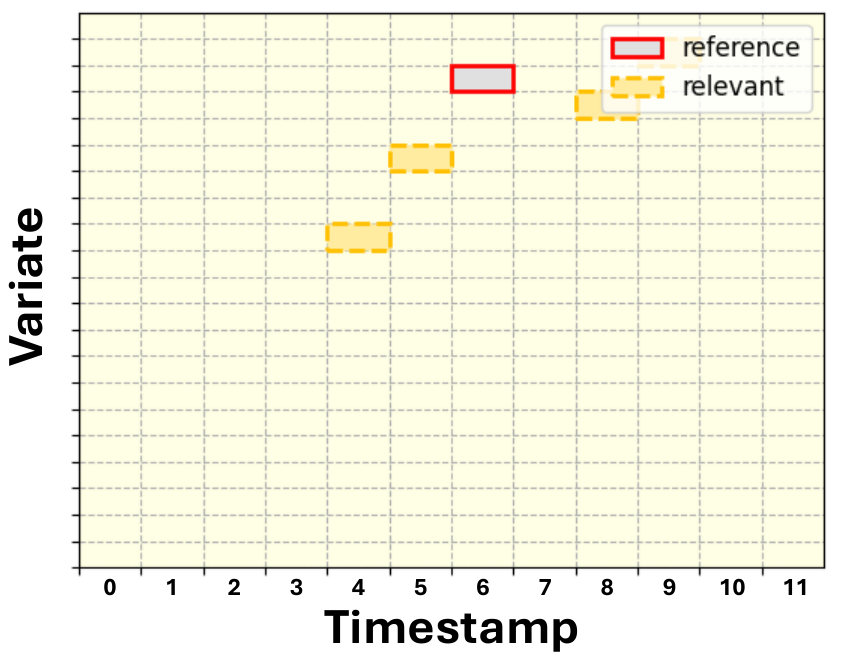}
    \includegraphics[width=0.33\textwidth]{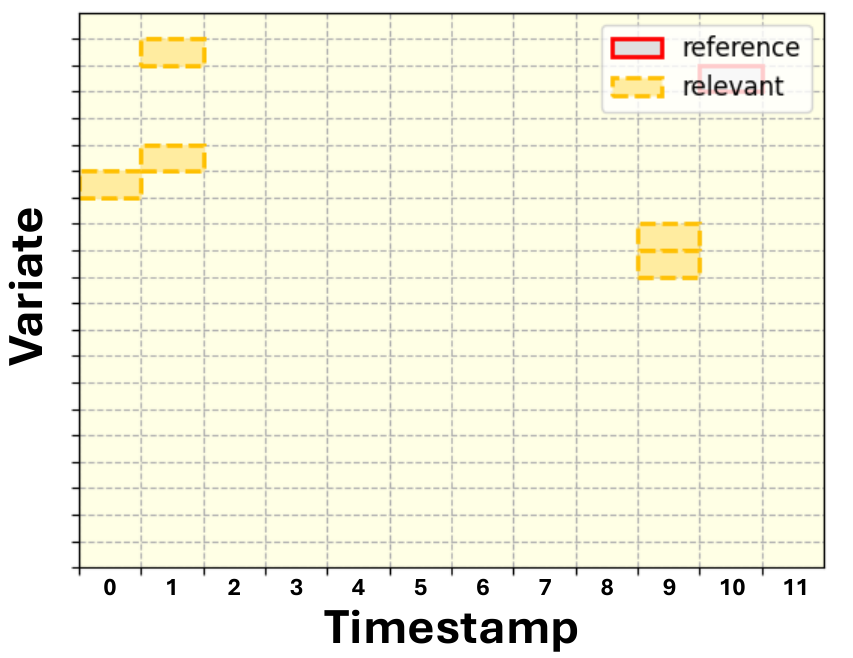}
    \caption{}
\end{subfigure}
\caption{Visualization of asynchronous interactions between the reference point and the relevant points from the Weather dataset.}
\label{fig:heatmap_weather}
\end{figure*}

To analyze the effectiveness of the JA attention module in capturing asynchronous interactions, we visualized the reference point $(t,v)$ along with the locations of the relevant features $R_q^{\text{cross}}-R_q^{\text{self}}$, excluding the self-axes, selected through DTV sampling, within the grid map.
The red box marks the reference point, and the yellow box highlights the features that are highly relevant to it across the temporal and variable dimensions.
Figs.~\ref{fig:heatmap_etth1},~\ref{fig:heatmap_exchange} and~\ref{fig:heatmap_weather} illustrate the possible asynchronous interactions in the MTS dataset, categorized into several scenarios and visualized using the ETTh1, Exchange, and Weather datasets, respectively.

% In each figure, (a) represents a scenario with many asynchronous interactions, while (b) represents a scenario with few asynchronous interactions, both under the same conditions of timestep and variate.
% (c) and (d) show how asynchronous interactions vary across variates within the same time window, depending on the reference point's variate.
% (e) and (f) illustrate the evolution of asynchronous dependencies within a single variate over time, with (f) depicting the relevant features one timestep after those in (e).
% As a result, we conclude from the visualizations in (a) to (e) that the locations of the selected relevant features vary depending on the reference point.

(a), (b), (c), and (d) in each figure illustrate how asynchronous interactions vary depending on the timestep of the reference point within a specific variate.
% As a result, we conclude that the visualizations in (a) to (d) demonstrate that the locations of the selected relevant features depend on the reference point.
As a result, these results suggest that the visualizations in (a) to (d) confirm that the locations of the selected relevant features are influenced by the reference point.

\section{Full Results}

Table~\ref{tbl:full_results_96_lscape} displays the full results of the multivariate long-term time series forecasting task, with the best and second-best results highlighted in \textbf{\textcolor{red}{red}} and \textcolor{blue}{blue}, respectively.
The results are reported for prediction lengths \(L_F \in \{96, 192, 336, 720\}\), using a fixed lookback window of \(L_H = 96\) for all baselines.
Avg represents the average value across the four prediction horizons.

\begin{figure}[!hbt]
\begin{subfigure}[t]{0.499\linewidth}
    \centering
    \includegraphics[width=\textwidth, height=3.3cm]{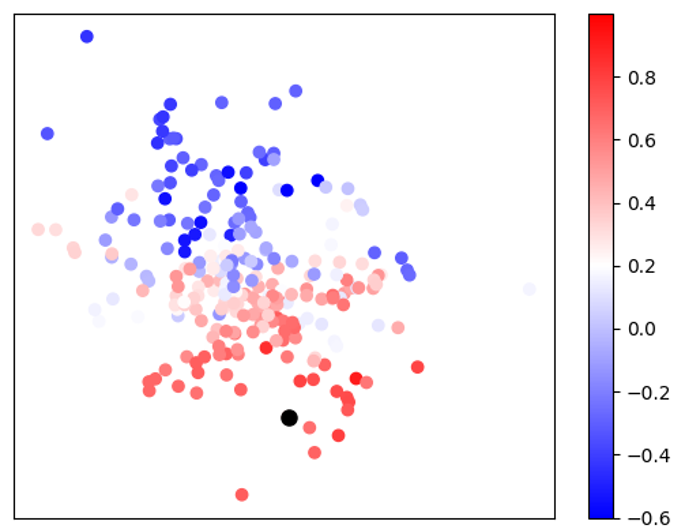}
\end{subfigure}
\hfill
\begin{subfigure}[t]{0.491\linewidth}
    \centering
    \includegraphics[width=0.99\textwidth, height=3.3cm]{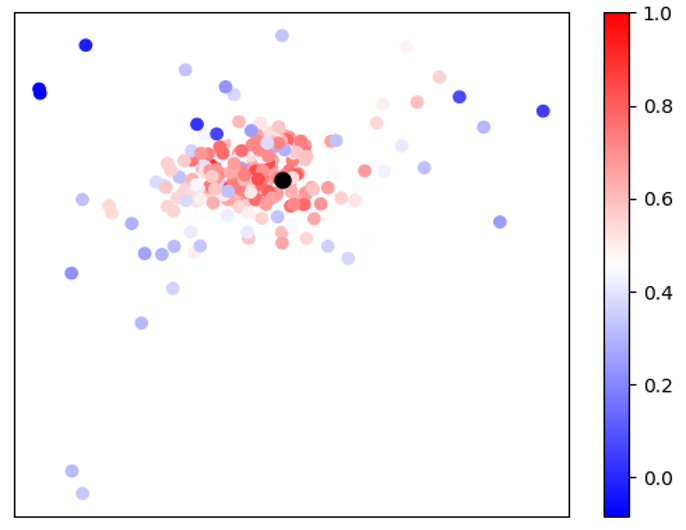}
\end{subfigure}
\begin{subfigure}[t]{0.499\linewidth}
    \centering
    \includegraphics[width=\textwidth]{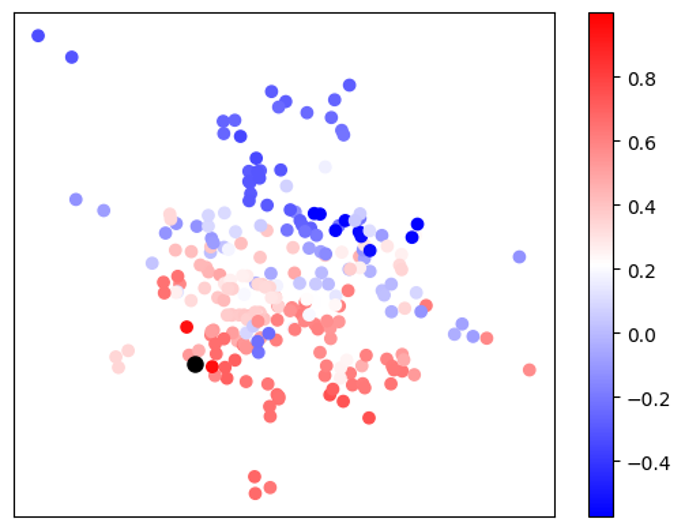}
\end{subfigure}
\hfill
\begin{subfigure}[t]{0.491\linewidth}
    \centering
    \includegraphics[width=\textwidth, height=3.3cm]{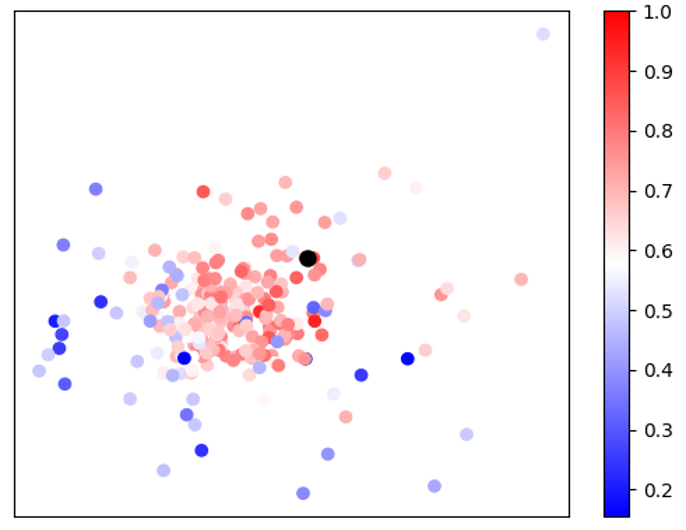}
\end{subfigure}
\begin{subfigure}[t]{0.499\linewidth}
    \centering
    \includegraphics[width=\textwidth]{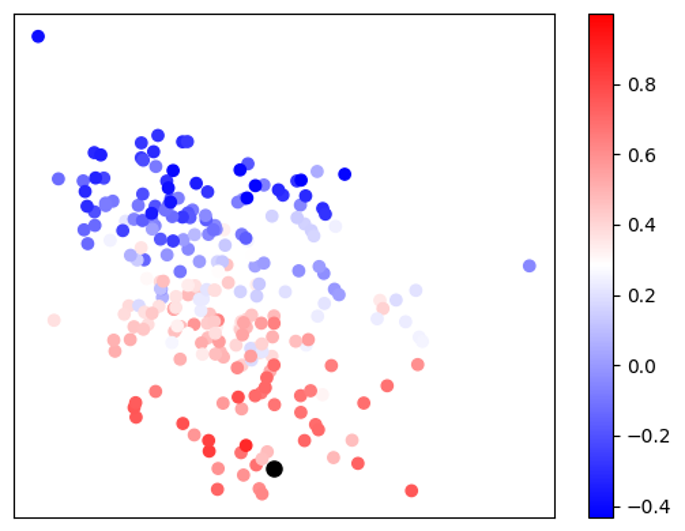}
\end{subfigure}
\hfill
\begin{subfigure}[t]{0.491\linewidth}
    \centering
    \includegraphics[width=\textwidth, height=3.3cm]{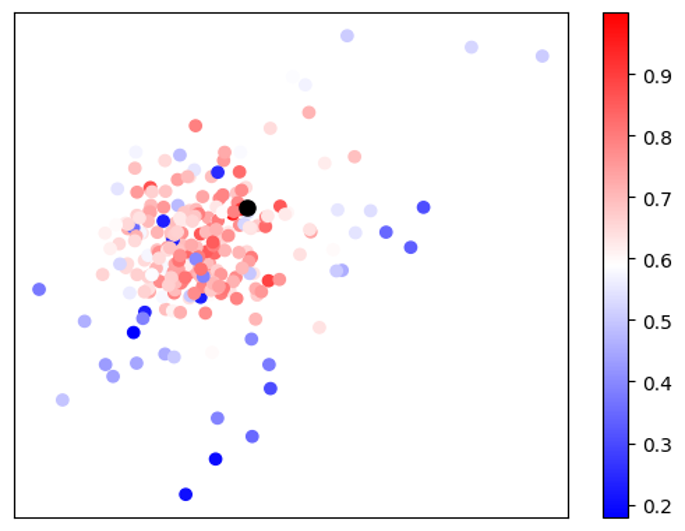}
\end{subfigure}
\hfill
\begin{subfigure}[t]{0.499\linewidth}
    \centering
    \includegraphics[width=\textwidth]{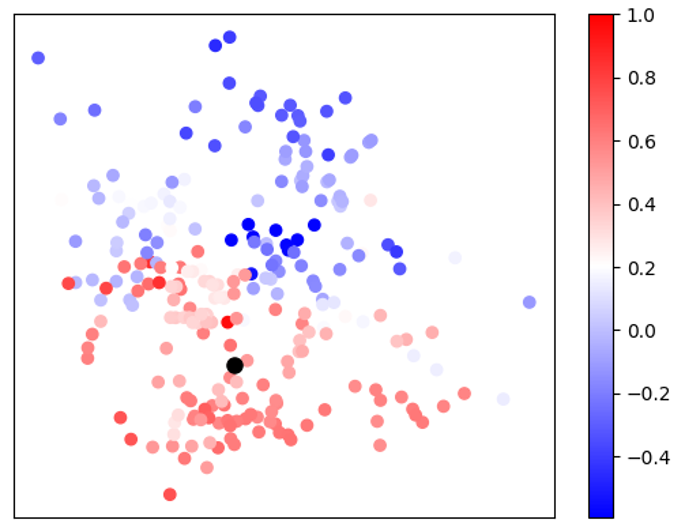}
\end{subfigure}
\hfill
\begin{subfigure}[t]{0.491\linewidth}
    \centering
    \includegraphics[width=\textwidth, height=3.3cm]{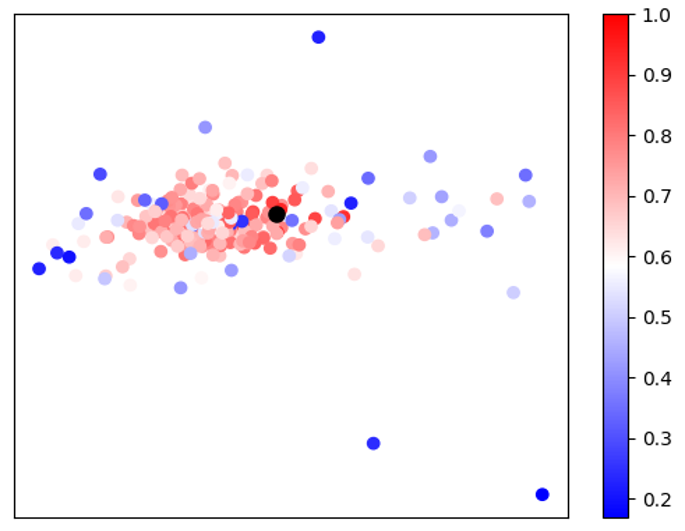}
\end{subfigure}
\hfill
\begin{subfigure}[t]{0.499\linewidth}
    \centering
    \includegraphics[width=\textwidth]{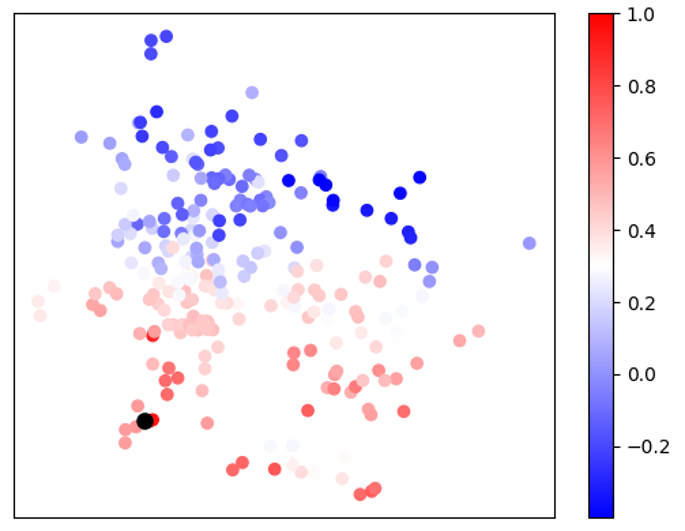}
    \caption{Trend}
    \label{fig5a}
\end{subfigure}
\hfill
\begin{subfigure}[t]{0.491\linewidth}
    \centering
    \includegraphics[width=\textwidth, height=3.3cm]{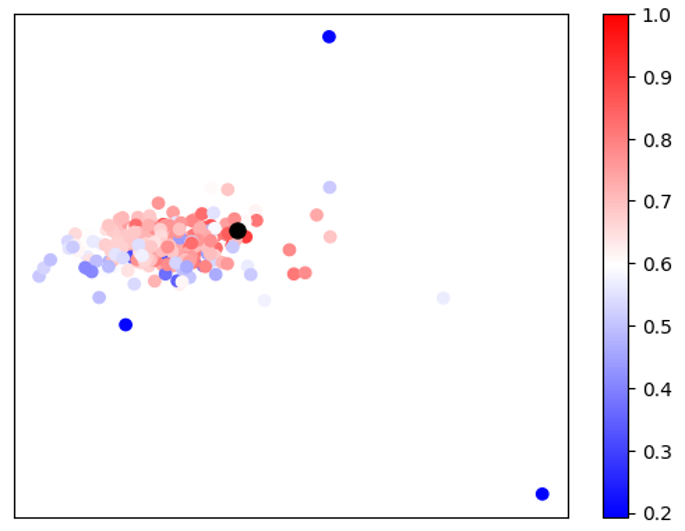}
    \caption{Seasonality}
    \label{fig5b}
\end{subfigure}
\caption{Qualitative analysis for DTV sampling is presented, with visualizations of five randomly selected reference points illustrating their spatial distribution within the 2D embedding space derived from the ETTh1 dataset.
The black points represent the query feature, while other features are colored based on their cosine similarity to the query: red for high similarity and blue for low similarity.
All 2D embedding spaces for the trend and seasonality components represent  $X^{Tr}$ and $X^{Se}$ for each paired input, respectively. 
}
\end{figure}

\input{table/appendix_full_results_96_lscape}

%% file: table/appendix_dataset.tex
% \begin{table}[h]
% \caption{Detailed descriptions of each dataset. 'Dim' represents the number of variables in each dataset. 'Dataset Size' indicates the total number of time points divided into (Train, Validation, Test) splits, respectively. 'Prediction Length' refers to the number of future time steps to forecast, with four prediction scenarios provided for each dataset. 'Frequency' specifies the time interval at which the data points are sampled.}
% \centering
% \small
% \renewcommand{\arraystretch}{1.3} % 행 간격 넓히기
% \begin{tabular}{c|c|c|c|c|c}
% \hline
% \textbf{Dataset} & \textbf{Dim} & \textbf{Prediction Length} & \textbf{Dataset Size} & \textbf{Frequency} & \textbf{Information} \\ \hline\hline
% ETTh1, ETTh2 & 7 & \{96, 192, 336, 720\} & (8545, 2881, 2881) & Hourly & Electricity \\ \hline
% ETTm1, ETTm2 & 7 & \{96, 192, 336, 720\} & (34465, 11521, 11521) & 15min & Electricity \\ \hline
% Exchange & 8 & \{96, 192, 336, 720\} & (5120, 665, 1422) & Daily & Economy \\ \hline
% Weather & 21 & \{96, 192, 336, 720\} & (36792, 5271, 10540) & 10min & Weather \\ \hline
% ECL & 321 & \{96, 192, 336, 720\} & (18317, 2633, 5261) & Hourly & Electricity \\ \hline
% Traffic & 862 & \{96, 192, 336, 720\} & (12185, 1757, 3509) & Hourly & Transportation \\ \hline
% PEMS04 & 307 & \{12, 24, 48, 96\} & (10172, 3375, 3375) & 5min & Transportation \\ \hline
% PEMS08 & 170 & \{12, 24, 48, 96\} & (10690, 3548, 3548) & 5min & Transportation \\ \hline

% \end{tabular}

% \label{tab:dataset_info}
% \end{table}

\begin{table}[h]
\resizebox{\columnwidth}{!}{%
\begin{tabular}{c|c|c|c}
\hline
Dataset & Dim & Dataset Size & Frequency \\
\hline
ETTh1, ETTh2 & 7 & (8545, 2881, 2881) & Hourly \\
ETTm1, ETTm2 & 7 & (34465, 11521, 11521) & 15 min \\
Exchange & 8 & (5120, 665, 1422) & Daily \\
Weather & 21 & (36792, 5271, 10540) & 10 min \\
ECL & 321 & (18317, 2633, 5261) & Hourly \\
Traffic & 862 & (12185, 1757, 3509) & Hourly \\
\hline
\end{tabular}
}
\caption{Dim represents the number of variates in each dataset, Dataset size indicates the total number of time points divided into training, validation, and test sets, and frequency specifies the interval between data samples.}
\label{tbl:dataset_info}
\end{table}

%% file: table/ablation_separate_sampling.tex
\begin{table}[t]
\resizebox{\columnwidth}{!}{%
\begin{tabular}{cll|cc|cc|cc}
\hline
\multicolumn{3}{l|}{\multirow{2}{*}{Applying DTV}} & \multicolumn{2}{c|}{ETTh1}      & \multicolumn{2}{c|}{Exchange}      & \multicolumn{2}{c}{Weather}     \\ \cline{4-9} 
\multicolumn{3}{l|}{}                                 & MSE            & MAE            & MSE            & MAE            & MSE            & MAE            \\ \hline
\multicolumn{3}{l|}{(a) Commonly}    & 0.463          & 0.445          &    0.371       & 0.409  & 0.244          & 0.274          \\ 
\multicolumn{3}{l|}{(b) Separately}  & \textbf{0.434} & \textbf{0.435} & \textbf{0.349}  & \textbf{0.398} & \textbf{0.240} & \textbf{0.270} \\ \hline
\end{tabular}
}
% \caption{Considering the inherent characteristics of time series data, an experiment was conducted to evaluate the effectiveness of the proposed separate sampling method. This is used for the purpose of further reflecting self-axis information based on TMS's analysis that information on the same time and same variable is more important.}
\caption{Ablation on separate sampling. (a) applies DTV sampling on the combined pool, while (b) performs separate sampling for the self-axis and cross-axis pools independently.}
\label{tbl:ablation_separate_sampling}
\end{table}

%% file: table/appendix_ablation_residual_conn.tex
\begin{table}[t]
\resizebox{\columnwidth}{!}{%
\begin{tabular}{c|cc|cc|cc}
\hline
\multicolumn{1}{c|}{\multirow{2}{*}{Method}} & \multicolumn{2}{c|}{ETTh1} & \multicolumn{2}{c|}{Exchange} & \multicolumn{2}{c}{Weather} \\ \cline{2-7} 
\multicolumn{1}{c|}{} & MSE & MAE & MSE & MAE & MSE & MAE \\ \hline
\multicolumn{1}{l|}{Decomposition} & 0.392 & 0.406 & 0.088 & 0.206 & \textbf{0.156} & 0.203 \\
\multicolumn{1}{l|}{+ Residual Connection } & \textbf{0.380} & \textbf{0.399} & \textbf{0.083} & \textbf{0.202} & \textbf{0.156} & \textbf{0.201} \\ \hline
\end{tabular}%
}
\caption{Ablation on residual connections in time series decomposition. Results are presented for a prediction length of $L_F = 96$ and a lookback length of $L_H = 96$.}
\label{tbl:residual_conn}
\end{table}

%% file: table/appendix_full_results_96_lscape.tex
% %Pleaseaddthefollowingrequiredpackagestoyourdocumentpreamble:
% \usepackage{multirow}
% \usepackage{graphicx}

\begin{landscape}
\begin{table}[p]
    \resizebox{\linewidth}{!}{
    \begin{tabular}{cc|cc|cc|cc|cc|cc|cc|cc|cc|cc|cc|cc|cc}

    \hline
    \multicolumn{2}{c|}{\multirow{2}{*}{Models}} & \multicolumn{2}{c|}{\textbf{Ours}} & \multicolumn{2}{c|}{TimeXer} & \multicolumn{2}{c|}{VCformer} & \multicolumn{2}{c|}{iTransformer} & \multicolumn{2}{c|}{TimeMixer} & \multicolumn{2}{c|}{DSformer} & \multicolumn{2}{c|}{PatchTST} & \multicolumn{2}{c|}{Crossformer} & \multicolumn{2}{c|}{TimesNet} & \multicolumn{2}{c|}{DLinear} & \multicolumn{2}{c|}{FEDformer} & \multicolumn{2}{c}{Autoformer} \\
    \multicolumn{2}{c|}{} & \multicolumn{2}{c|}{\textbf{(2025)}} & \multicolumn{2}{c|}{(2024)} & \multicolumn{2}{c|}{(2024)} & \multicolumn{2}{c|}{(2024)} & \multicolumn{2}{c|}{(2024)} & \multicolumn{2}{c|}{(2023)} & \multicolumn{2}{c|}{(2023)}& \multicolumn{2}{c|}{(2023)} & \multicolumn{2}{c|}{(2023)} & \multicolumn{2}{c|}{(2023)} & \multicolumn{2}{c|}{(2022)} & \multicolumn{2}{c}{(2021)} \\ \midrule
    \multicolumn{2}{c|}{Metrics} & MSE & \multicolumn{1}{c|}{MAE} & MSE & \multicolumn{1}{c|}{MAE} & MSE & \multicolumn{1}{c|}{MAE} & MSE & \multicolumn{1}{c|}{MAE} & MSE & \multicolumn{1}{c|}{MAE} & MSE & \multicolumn{1}{c|}{MAE} & MSE & \multicolumn{1}{c|}{MAE} & MSE & \multicolumn{1}{c|}{MAE} & MSE & \multicolumn{1}{c|}{MAE} & MSE & \multicolumn{1}{c|}{MAE} & MSE & \multicolumn{1}{c|}{MAE} & MSE & MAE \\ \hline

    \multicolumn{1}{c|}{\multirow{5}{*}{\rotatebox{90}{ETTh1}}}&96&0.380&\textcolor{blue}{0.399}&0.382&0.403&0.376&\textbf{\textcolor{red}{0.397}}&0.386&0.405&\textcolor{blue}{0.395}&0.400&\textbf{\textcolor{red}{0.373}}&\textbf{\textcolor{red}{0.397}}&0.414&0.419&0.423&0.448&0.384&0.402&0.386&0.400&0.376&0.419&0.449&0.459\\
    \multicolumn{1}{c|}{}&192&0.425&0.431&0.429&0.435&0.431&0.427&0.441&0.436&0.429&\textbf{\textcolor{red}{0.421}}&\textbf{\textcolor{red}{0.419}}&\textcolor{blue}{0.425}&0.460&0.445&0.471&0.474&0.436&0.429&0.437&0.432&\textcolor{blue}{0.420}&0.448&0.500&0.482\\
    \multicolumn{1}{c|}{}&336&0.469&0.450&0.468&\textcolor{blue}{0.448}&0.473&0.449&0.487&0.458&0.484&0.458&\textbf{\textcolor{red}{0.457}}&\textbf{\textcolor{red}{0.446}}&0.501&0.466&0.570&0.546&0.491&0.469&0.481&0.459&\textcolor{blue}{0.459}&0.465&0.521&0.496\\
    \multicolumn{1}{c|}{}&720&\textbf{\textcolor{red}{0.463}}&\textbf{\textcolor{red}{0.460}}&\textcolor{blue}{0.469}&\textcolor{blue}{0.461}&0.476&0.474&0.503&0.491&0.498&0.482&0.499&0.497&0.500&0.488&0.653&0.621&0.521&0.500&0.519&0.516&0.506&0.507&0.514&0.512\\\cline{2-26}
    \multicolumn{1}{c|}{}&Avg&\textbf{\textcolor{red}{0.434}}&\textbf{\textcolor{red}{0.435}}&\textcolor{blue}{0.437}&\textcolor{blue}{0.437}&0.439&\textcolor{blue}{0.437}&0.454&0.447&0.447&0.440&\textcolor{blue}{0.437}&0.441&0.469&0.454&0.529&0.522&0.458&0.450&0.456&0.452&0.440&0.460&0.496&0.487\\\hline

    \multicolumn{1}{c|}{\multirow{5}{*}{\rotatebox{90}{ETTh2}}}&96&0.290&\textcolor{blue}{0.340}&\textbf{\textcolor{red}{0.286}}&\textbf{\textcolor{red}{0.338}}&0.292&0.344&0.297&0.349&\textcolor{blue}{0.289}&0.341&0.296&0.351&0.302&0.348&0.745&0.584&0.340&0.374&0.333&0.387&0.358&0.397&0.346&0.388\\
    \multicolumn{1}{c|}{}&192&\textcolor{blue}{0.372}&0.394&\textbf{\textcolor{red}{0.363}}&\textbf{\textcolor{red}{0.389}}&0.377&0.396&0.380&0.400&\textcolor{blue}{0.372}&\textcolor{blue}{0.392}&0.399&0.414&0.388&0.400&0.877&0.656&0.402&0.414&0.477&0.476&0.429&0.439&0.456&0.452\\
    \multicolumn{1}{c|}{}&336&\textcolor{blue}{0.400}&\textcolor{blue}{0.421}&0.414&0.423&0.417&0.430&0.428&0.432&\textbf{\textcolor{red}{0.386}}&\textbf{\textcolor{red}{0.414}}&0.434&0.443&0.426&0.433&1.043&0.731&0.452&0.452&0.594&0.541&0.496&0.487&0.482&0.486\\
    \multicolumn{1}{c|}{}&720&0.416&0.446&\textbf{\textcolor{red}{0.408}}&\textbf{\textcolor{red}{0.432}}&0.423&0.443&0.427&0.445&\textcolor{blue}{0.412}&\textcolor{blue}{0.434}&0.454&0.463&0.431&0.446&1.104&0.763&0.462&0.468&0.831&0.657&0.463&0.474&0.515&0.511\\\cline{2-26}
    \multicolumn{1}{c|}{}&Avg&0.370&0.400&\textcolor{blue}{0.367}&\textcolor{blue}{0.396}&0.377&0.403&0.383&0.407&\textbf{\textcolor{red}{0.364}}&\textbf{\textcolor{red}{0.395}}&0.396&0.418&0.387&0.407&0.942&0.684&0.414&0.427&0.559&0.515&0.437&0.449&0.450&0.459\\\hline

    \multicolumn{1}{c|}{\multirow{5}{*}{\rotatebox{90}{ETTm1}}}&96&\textbf{\textcolor{red}{0.314}}&\textbf{\textcolor{red}{0.354}}&\textcolor{blue}{0.318}&\textcolor{blue}{0.356}&0.319&0.359&0.334&0.368&0.320&0.357&0.326&0.364&0.329&0.367&0.404&0.426&0.338&0.375&0.345&0.372&0.379&0.419&0.505&0.475\\
    \multicolumn{1}{c|}{}&192&\textcolor{blue}{0.361}&0.383&0.362&0.383&0.364&\textcolor{blue}{0.382}&0.377&0.391&0.361&\textbf{\textcolor{red}{0.381}}&\textbf{\textcolor{red}{0.360}}&\textcolor{blue}{0.382}&0.367&0.385&0.450&0.451&0.374&0.387&0.380&0.389&0.426&0.441&0.553&0.496\\
    \multicolumn{1}{c|}{}&336&\textbf{\textcolor{red}{0.390}}&\textcolor{blue}{0.405}&0.395&0.407&0.399&\textcolor{blue}{0.405}&0.426&0.420&\textbf{\textcolor{red}{0.390}}&\textbf{\textcolor{red}{0.404}}&\textcolor{blue}{0.394}&\textcolor{blue}{0.405}&0.399&0.410&0.532&0.515&0.410&0.411&0.413&0.413&0.445&0.459&0.621&0.537\\
    \multicolumn{1}{c|}{}&720&0.455&0.446&\textbf{\textcolor{red}{0.452}}&\textcolor{blue}{0.441}&0.467&0.442&0.491&0.459&\textcolor{blue}{0.454}&\textcolor{blue}{0.441}&0.474&0.451&\textcolor{blue}{0.454}&\textbf{\textcolor{red}{0.439}}&0.666&0.589&0.478&0.450&0.474&0.453&0.543&0.490&0.671&0.561\\\cline{2-26}
    \multicolumn{1}{c|}{}&Avg&\textbf{\textcolor{red}{0.380}}&\textcolor{blue}{0.397}&0.382&\textcolor{blue}{0.397}&0.387&\textcolor{blue}{0.397}&0.407&0.410&\textcolor{blue}{0.381}&\textbf{\textcolor{red}{0.395}}&0.389&0.401&0.387&0.400&0.513&0.496&0.400&0.406&0.403&0.407&0.448&0.452&0.588&0.517\\\hline

    \multicolumn{1}{c|}{\multirow{5}{*}{\rotatebox{90}{ETTm2}}}&96&\textcolor{blue}{0.173}&\textcolor{blue}{0.258}&\textbf{\textcolor{red}{0.171}}&\textbf{\textcolor{red}{0.256}}&0.180&0.266&0.180&0.264&0.175&\textcolor{blue}{0.258}&0.201&0.286&0.175&0.259&0.287&0.366&0.187&0.267&0.193&0.292&0.203&0.287&0.255&0.339\\
    \multicolumn{1}{c|}{}&192&\textcolor{blue}{0.238}&0.303&\textbf{\textcolor{red}{0.237}}&\textbf{\textcolor{red}{0.299}}&0.245&0.306&0.250&0.309&\textbf{\textcolor{red}{0.237}}&\textbf{\textcolor{red}{0.299}}&0.281&0.335&0.241&\textcolor{blue}{0.302}&0.414&0.492&0.249&0.309&0.284&0.362&0.269&0.328&0.281&0.340\\
    \multicolumn{1}{c|}{}&336&0.299&\textcolor{blue}{0.339}&\textbf{\textcolor{red}{0.296}}&\textbf{\textcolor{red}{0.338}}&0.307&0.345&0.311&0.348&\textcolor{blue}{0.298}&0.340&0.336&0.367&0.305&0.343&0.597&0.542&0.321&0.351&0.369&0.427&0.325&0.366&0.339&0.372\\
    \multicolumn{1}{c|}{}&720&0.395&0.398&\textcolor{blue}{0.392}&\textbf{\textcolor{red}{0.394}}&0.406&0.402&0.412&0.407&\textbf{\textcolor{red}{0.391}}&\textcolor{blue}{0.396}&0.430&0.417&0.402&0.400&1.730&1.042&0.408&0.403&0.554&0.522&0.421&0.415&0.433&0.432\\\cline{2-26}
    \multicolumn{1}{c|}{}&Avg&0.276&0.325&\textbf{\textcolor{red}{0.274}}&\textbf{\textcolor{red}{0.322}}&0.285&0.330&0.288&0.332&\textcolor{blue}{0.275}&\textcolor{blue}{0.323}&0.312&0.351&0.281&0.326&0.757&0.610&0.291&0.333&0.350&0.401&0.305&0.349&0.327&0.371\\\hline

    \multicolumn{1}{c|}{\multirow{5}{*}{\rotatebox{90}{Exchange}}}&96&\textbf{\textcolor{red}{0.083}}&\textbf{\textcolor{red}{0.201}}&0.244&0.209&\textcolor{blue}{0.085}&0.205&0.086&0.206&\textbf{\textcolor{red}{0.083}}&\textcolor{blue}{0.204}&0.092&0.216&0.088&0.205&0.256&0.367&0.107&0.234&0.088&0.218&0.148&0.278&0.197&0.323\\
    \multicolumn{1}{c|}{}&192&\textbf{\textcolor{red}{0.174}}&\textbf{\textcolor{red}{0.297}}&0.192&0.311&\textcolor{blue}{0.176}&\textcolor{blue}{0.299}&0.177&\textcolor{blue}{0.299}&0.182&0.304&0.189&0.312&\textcolor{blue}{0.176}&\textcolor{blue}{0.299}&0.470&0.509&0.226&0.344&\textcolor{blue}{0.176}&0.315&0.271&0.315&0.300&0.369\\
    \multicolumn{1}{c|}{}&336&0.336&0.417&0.363&0.435&0.328&\textcolor{blue}{0.415}&0.331&0.417&0.361&0.437&0.348&0.430&\textbf{\textcolor{red}{0.301}}&\textbf{\textcolor{red}{0.397}}&1.268&0.883&0.367&0.448&\textcolor{blue}{0.313}&0.427&0.460&0.427&0.509&0.524\\
    \multicolumn{1}{c|}{}&720&\textbf{\textcolor{red}{0.800}}&\textbf{\textcolor{red}{0.674}}&0.888&0.711&\textcolor{blue}{0.830}&\textcolor{blue}{0.688}&0.847&0.697&0.963&0.710&0.947&0.740&0.901&0.714&1.767&1.068&0.964&0.746&0.839&0.695&1.195&0.695&1.447&0.941\\\cline{2-26}
    \multicolumn{1}{c|}{}&Avg&\textbf{\textcolor{red}{0.349}}&\textbf{\textcolor{red}{0.398}}&0.422&0.416&0.355&\textcolor{blue}{0.402}&0.360&0.403&0.397&0.414&0.394&0.425&0.367&0.404&0.940&0.707&0.416&0.443&\textcolor{blue}{0.353}&0.414&0.519&0.429&0.613&0.539\\\hline

    \multicolumn{1}{c|}{\multirow{5}{*}{\rotatebox{90}{Weather}}}&96&\textbf{\textcolor{red}{0.156}}&\textbf{\textcolor{red}{0.201}}&\textcolor{blue}{0.157}&\textcolor{blue}{0.205}&0.171&0.220&0.174&0.214&0.163&0.209&0.170&0.217&0.177&0.218&0.158&0.230&0.172&0.220&0.196&0.255&0.217&0.296&0.266&0.336\\
    \multicolumn{1}{c|}{}&192&\textbf{\textcolor{red}{0.203}}&\textbf{\textcolor{red}{0.247}}&\textcolor{blue}{0.204}&\textbf{\textcolor{red}{0.247}}&0.230&0.266&0.221&0.254&0.208&\textcolor{blue}{0.250}&0.253&0.296&0.219&0.261&0.206&0.277&0.240&0.271&0.237&0.296&0.276&0.336&0.307&0.367\\
    \multicolumn{1}{c|}{}&336&\textcolor{blue}{0.260}&0.291&0.261&\textcolor{blue}{0.290}&0.280&0.299&0.278&0.296&\textbf{\textcolor{red}{0.251}}&\textbf{\textcolor{red}{0.287}}&0.285&0.310&0.278&0.297&0.272&0.335&0.280&0.306&0.283&0.335&0.339&0.380&0.359&0.395\\
    \multicolumn{1}{c|}{}&720&0.341&\textcolor{blue}{0.343}&\textcolor{blue}{0.340}&\textbf{\textcolor{red}{0.341}}&0.352&0.344&0.358&0.347&\textbf{\textcolor{red}{0.339}}&\textbf{\textcolor{red}{0.341}}&0.395&0.391&0.354&0.348&0.398&0.418&0.365&0.359&0.345&0.381&0.403&0.428&0.419&0.428\\\cline{2-26}
    \multicolumn{1}{c|}{}&Avg&\textbf{\textcolor{red}{0.240}}&\textbf{\textcolor{red}{0.270}}&\textcolor{blue}{0.241}&\textcolor{blue}{0.271}&0.258&0.282&0.258&0.278&\textbf{\textcolor{red}{0.240}}&\textcolor{blue}{0.271}&0.276&0.304&0.259&0.281&0.259&0.315&0.259&0.287&0.265&0.317&0.309&0.360&0.338&0.382\\\hline

    \multicolumn{1}{c|}{\multirow{5}{*}{\rotatebox{90}{Electricity}}}&96&\textbf{\textcolor{red}{0.136}}&\textbf{\textcolor{red}{0.232}}&\textcolor{blue}{0.140}&0.242&0.150&0.242&0.148&\textcolor{blue}{0.240}&0.153&0.247&0.164&0.261&0.181&0.270&0.219&0.314&0.168&0.272&0.197&0.282&0.193&0.308&0.201&0.317\\
    \multicolumn{1}{c|}{}&192&\textbf{\textcolor{red}{0.157}}&\textbf{\textcolor{red}{0.253}}&\textbf{\textcolor{red}{0.157}}&0.256&0.167&\textcolor{blue}{0.255}&\textcolor{blue}{0.162}&\textbf{\textcolor{red}{0.253}}&0.166&0.256&0.177&0.272&0.188&0.274&0.231&0.322&0.184&0.289&0.196&0.285&0.201&0.315&0.222&0.334\\
    \multicolumn{1}{c|}{}&336&\textbf{\textcolor{red}{0.174}}&0.271&\textcolor{blue}{0.176}&0.275&0.182&\textcolor{blue}{0.270}&0.178&\textbf{\textcolor{red}{0.269}}&0.185&0.277&0.201&0.294&0.204&0.293&0.246&0.337&0.198&0.300&0.209&0.301&0.214&0.329&0.231&0.338\\
    \multicolumn{1}{c|}{}&720&\textbf{\textcolor{red}{0.197}}&\textbf{\textcolor{red}{0.292}}&\textcolor{blue}{0.211}&0.306&0.221&\textcolor{blue}{0.302}&0.225&0.317&0.225&0.310&0.242&0.327&0.246&0.324&0.280&0.363&0.220&0.320&0.245&0.333&0.246&0.355&0.254&0.361\\\cline{2-26}
    \multicolumn{1}{c|}{}&Avg&\textbf{\textcolor{red}{0.166}}&\textbf{\textcolor{red}{0.262}}&\textcolor{blue}{0.171}&0.270&0.180&\textcolor{blue}{0.267}&0.178&0.270&0.182&0.272&0.196&0.289&0.205&0.290&0.244&0.334&0.192&0.295&0.212&0.300&0.214&0.327&0.227&0.338\\\hline

    \multicolumn{1}{c|}{\multirow{5}{*}{\rotatebox{90}{Traffic}}}&96&\textcolor{blue}{0.408}&0.286&0.428&\textcolor{blue}{0.271}&0.454&0.310&\textbf{\textcolor{red}{0.395}}&\textbf{\textcolor{red}{0.268}}&0.462&0.285&0.546&0.352&0.462&0.295&0.522&0.290&0.593&0.321&0.650&0.396&0.587&0.366&0.613&0.388\\
    \multicolumn{1}{c|}{}&192&\textcolor{blue}{0.427}&0.290&0.448&\textcolor{blue}{0.282}&0.468&0.315&\textbf{\textcolor{red}{0.417}}&\textbf{\textcolor{red}{0.276}}&0.473&0.296&0.547&0.347&0.466&0.296&0.530&0.293&0.617&0.336&0.598&0.370&0.604&0.373&0.616&0.382\\
    \multicolumn{1}{c|}{}&336&\textcolor{blue}{0.441}&0.298&0.473&\textcolor{blue}{0.289}&0.486&0.325&\textbf{\textcolor{red}{0.433}}&\textbf{\textcolor{red}{0.283}}&0.498&0.296&0.562&0.352&0.482&0.304&0.558&0.305&0.629&0.336&0.605&0.373&0.621&0.383&0.622&0.337\\
    \multicolumn{1}{c|}{}&720&\textcolor{blue}{0.473}&0.315&0.516&\textcolor{blue}{0.307}&0.524&0.348&\textbf{\textcolor{red}{0.467}}&\textbf{\textcolor{red}{0.302}}&0.506&0.313&0.597&0.370&0.514&0.322&0.589&0.328&0.640&0.350&0.645&0.394&0.626&0.382&0.660&0.408\\\cline{2-26}
    \multicolumn{1}{c|}{}&Avg&\textcolor{blue}{0.437}&0.297&0.466&\textcolor{blue}{0.287}&0.483&0.325&\textbf{\textcolor{red}{0.428}}&\textbf{\textcolor{red}{0.282}}&0.484&0.297&0.563&0.355&0.481&0.304&0.550&0.304&0.620&0.336&0.625&0.383&0.610&0.376&0.628&0.379\\\hline

    \end{tabular}
    }
\centering
\caption{Full results of the multivariate long-term time series forecasting task.}
\label{tbl:full_results_96_lscape}
\end{table}
\end{landscape}